%% file: main.tex
\documentclass[10pt,twocolumn,letterpaper]{article}

\usepackage[pagenumbers]{cvpr} 


\definecolor{cvprblue}{rgb}{0.21,0.49,0.74}
\definecolor{mygreen}{HTML}{009901}
\usepackage[pagebackref,breaklinks,colorlinks,allcolors=cvprblue]{hyperref}

\usepackage{multirow}
\usepackage{multicol}
\usepackage{stfloats}
\usepackage{xcolor,colortbl}
\newcommand{\R}{\mathbb{R}}
\usepackage{xcolor}
\usepackage{comment}
\usepackage[flushleft]{threeparttable}
\usepackage{tcolorbox}
\usepackage{algorithm,algorithmicx,algpseudocode}
\usepackage[framemethod=tikz]{mdframed}
\gdef\Sepline{%
  \par\noindent\makebox[\linewidth][l]{%
  \hspace*{-\mdflength{innerleftmargin}}%
   \tikz\draw[thick,dashed,gray!60] (0,0) --%
(\textwidth+\the\mdflength{innerleftmargin}+\the\mdflength{innerrightmargin},0);
  }\par\nobreak}

\title{SnapGen: Taming High-Resolution Text-to-Image Models for Mobile Devices \\ with Efficient Architectures and Training}

\author{
Dongting Hu\textsuperscript{1,2,*} Jierun Chen\textsuperscript{1,3,*} Xijie Huang\textsuperscript{1,3,*}
Huseyin Coskun\textsuperscript{1}  Arpit Sahni\textsuperscript{1} \\
Aarush Gupta\textsuperscript{1}   Anujraaj Goyal\textsuperscript{1}  Dishani Lahiri\textsuperscript{1}   Rajesh Singh\textsuperscript{1}
Yerlan Idelbayev\textsuperscript{1} \\
Junli Cao\textsuperscript{1}  Yanyu Li\textsuperscript{1} Kwang-Ting Cheng\textsuperscript{3} S.-H. Gary Chan\textsuperscript{3}   Mingming Gong\textsuperscript{2, 4} \\
Sergey Tulyakov\textsuperscript{1}  Anil Kag\textsuperscript{1,$\dagger$} Yanwu Xu\textsuperscript{1,$\dagger$} Jian Ren\textsuperscript{1,$\dagger$} \\
\vspace{0.001em}\\
\textsuperscript{1}~Snap Inc.\quad \textsuperscript{2}~The University of Melbourne \quad \textsuperscript{3}~HKUST \quad \textsuperscript{4}~MBZUAI\\
\textsuperscript{*}~Equal contribution \quad \textsuperscript{$\dagger$}~Equal advising \\
Project Page: \href{https://snap-research.github.io/snapgen}{https://snap-research.github.io/snapgen}
\vspace{-3ex}
}
\begin{document}

\twocolumn[{
\renewcommand\twocolumn[1][]{#1}
\maketitle
\begin{center}
    \centering
    \captionsetup{type=figure}
    \includegraphics[width=0.92\textwidth]{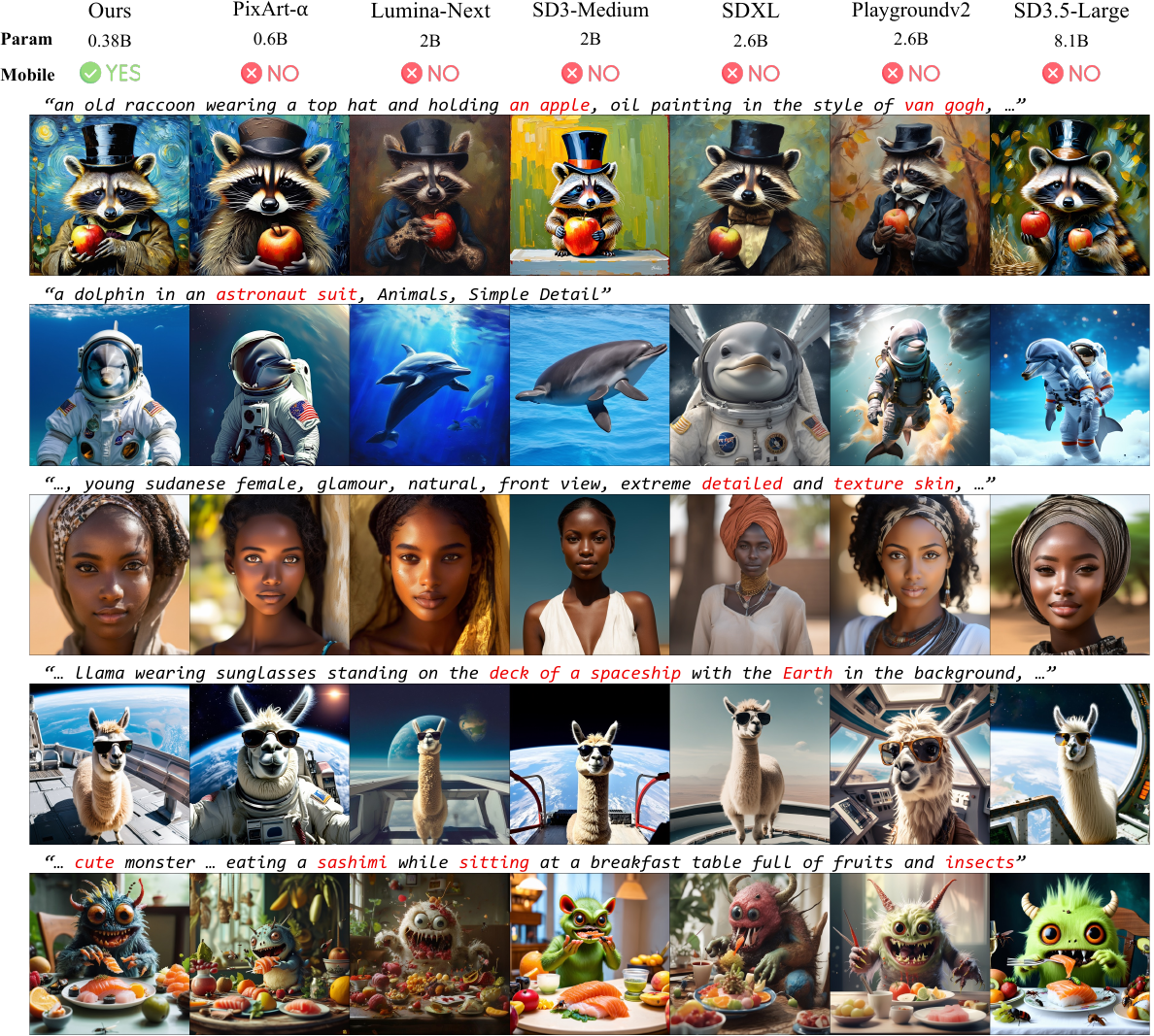}
    \caption{Comparison of various text-to-image models in terms of model size, mobile device compatibility, and visual output quality. Our model, with only \emph{379M} parameters, demonstrates competitive visual quality while being mobile-compatible.  Input text prompts are shown above each image grid; all images are generated at $1024^2$ resolution—zoom in for details. More examples are shown in \href{https://snap-research.github.io/snapgen}{webpage}.}{
    }
    \label{fig:teaser}
\end{center}
}]

\input{sec/0_abstract}    
\input{sec/1_intro}
\input{sec/2_related_work}
\input{sec/3_method}
\input{sec/4_experiments_v2}

\input{sec/5_conclusion}
{
    \small
    \bibliographystyle{ieeenat_fullname}
    \bibliography{main}
}

\input{sec/X_suppl}

\end{document}

%% file: sec/0_abstract.tex
\begin{abstract}
Existing text-to-image (T2I) diffusion models face several limitations, including large model sizes, slow runtime, and low-quality generation on mobile devices. This paper aims to address all of these challenges by developing \textbf{an extremely small and fast T2I model that generates high-resolution and high-quality images on mobile platforms}. We propose several techniques to achieve this goal. First, we systematically examine the design choices of the network architecture to reduce model parameters and latency, while ensuring high-quality generation. Second, to further improve generation quality, we employ cross-architecture knowledge distillation from a much larger model, using a multi-level approach to guide the training of our model from scratch. Third, we enable a few-step generation by integrating adversarial guidance with knowledge distillation.
For the first time, our model SnapGen, demonstrates the generation of $1024^2$ px images on a mobile device around $1.4$ seconds. On ImageNet-1K, our model, with only $372$M parameters, achieves an FID of $2.06$ for $256^2$ px generation. On T2I benchmarks (\ie, GenEval and DPG-Bench), our model with merely $379$M parameters, surpasses large-scale models with billions of parameters at a significantly smaller size (\eg, $7\times$ smaller than SDXL, $14\times$ smaller than IF-XL).
\end{abstract}

%% file: sec/1_intro.tex
\section{Introduction}
\label{sec:intro}

Large-scale text-to-image (T2I) diffusion models~\cite{podell2023sdxl,rombach2022high,deepfloyd,dalle_model,dalle2_model,imagen_model,chen2023pixart,chen2024pixart} have achieved remarkable success in content generation, powering numerous applications like image editing~\cite{zhang2023sine,liu2023hyperhuman,xiao2024omnigen,sheynin2024emu} and video creation~\cite{yang2024cogvideox,menapace2024snap,polyak2024movie}. However, T2I models often come with substantial model sizes and slow runtime, and deploying them on the cloud raises concerns related to data security and high costs~\cite{song2024sdxs}.

To address these challenges, there is huge growing interest in developing smaller and faster T2I models through techniques like model compression (\eg, pruning and quantization)~\cite{li2024snapfusion,zhao2023mobilediffusion,sui2024bitsfusion}, step reduction by distillation~\cite{xu2024ufogen,yin2024one}, and efficient attention mechanisms that mitigate the quadratic complexity~\cite{liu2024linfusion,sana}. Nevertheless, current works still encounter limitations, \eg, low-resolution generation on mobile devices, that constrain their broader application.

Most importantly, a critical question remains unexplored: \emph{how can we train a T2I model from scratch to generate high-quality, high-resolution images on mobile?} Such a model would offer substantial advantages in speed, compactness, cost-effectiveness, and secure deployment. To build this model, we introduce several innovations:
\begin{itemize}
    \item \textbf{Efficient Network Architectures:}
    We conduct an in-depth examination of network architectures, including the denoising \emph{UNet} and \emph{Autoencoder} (AE),  to obtain optimal trade-off between resource usage and performance. Unlike prior works that optimize and compress pre-trained diffusion models~\cite{castells2024ld,zhang2024laptop,kim2023bk}, we directly focus on macro- and micro-level design choices to achieve a novel architecture that greatly reduces model size and computational complexity, while preserving high-quality generation.
        
    \item \textbf{Improved Training Techniques:} We introduce several improvements to train a \emph{compact} T2I model from \emph{scratch}.
    We utilize flow matching~\cite{lipman2022flow,liu2022flow} as objective, aligning with larger models like SD3~\cite{esser2024scaling} and SD3.5~\cite{sd3.5}. This design enables effective knowledge and step distillation, transferring rich representations from large-scale diffusion models to our much smaller one.
    Further, we propose a multi-level knowledge distillation with a timestep-aware scaling that combines multiple training objectives. Instead of weighting objectives through a linear combination as in prior works~\cite{kim2023bk,liu2024linfusion}, we consider \emph{target prediction difficulty} (i.e., student-teacher difference) across various timesteps in flow matching.


    \item \textbf{Advanced Step Distillation:} We perform step distillation on our model by combining the adversarial training along with the knowledge distillation using a few-step teacher model (\ie, SD3.5-Large-Turbo~\cite{sd3.5-large-turbo}), enabling ultra-fast high-quality generation with only $4$ or $8$ steps.

\end{itemize}

\noindent We demonstrate the superior advantages of our approach and model through extensive experiments: 
\begin{itemize}
    \item On ImageNet-1K~\cite{deng2009imagenet} class-conditional image generation task, our model achieves the FID comparable to existing works with significantly reduced model size and computation, \ie, 
     half the model size and one-third of the compute resources compared with SiT-XL\citep{ma2024sit}, as in ~\cref{tab:imagenet_comparison}. 
    \item For large-scale T2I generation, our UNet model, with only $379$M parameters, demonstrates superior generation quality compared to billion-parameter models~\cite{podell2023sdxl,lin2024sdxllightning, zhuo2024lumina}, \eg, improved metrics on benchmark datasets 
    (\cref{tab:quantitative_benchmarks}) 
    and human evaluation (\cref{fig:human_evaluation}).
    \item Our compressed decoder, trained from scratch, has competitive reconstruction quality compared to commonly used models~\cite{podell2023sdxl, esser2024scaling}, with more than $36\times$ smaller size, enabling the mobile deployment.
    \item Notably, for the \emph{first} time, we show a T2I model achieving high-resolution generation (\eg, $1024^2$ px) on mobile devices (\eg, iPhone 16 Pro-Max) around $1.4$ seconds.
\end{itemize}

%% file: sec/2_related_work.tex
\section{Related Work}
\label{sec:related_work}
\noindent\textbf{High-Resolution Text-to-Image Models} have emerged with advanced architectures and multi-stage approaches designed to enhance visual fidelity and user customization. 
SDXL~\citep{podell2023sdxl} is a pioneering work in this field, employing a refined cascading approach with UNet backbone to generate high-detailed images, resulting in photorealistic outputs that maintain sharpness and clarity. 
The following studies explore different techniques like more advanced text encoders, better image refinement, or improved dataset preparation, to obtain better text-image alignment or higher-quality generation~\cite{deepfloyd,playground-v1,playground-v2,li2024playground,liu2024playground,li2024hunyuan,kandinsky2.1,arkhipkin2024kandinsky,liu2023hyperhuman,kolors,gao2024lumina,kag2024ascan}. However, most of these models contain billions of parameters, making them extremely slow, and not being able to run on resource-contained hardware like mobile devices. In this work, we aim to build a small and fast model that can perform high-resolution generation even on mobile platforms.

\noindent\textbf{Efficient Diffusion Models} address the challenges of bulky model size and long runtime. There have been efforts exploring the architecture optimization to remove redundancy from large models, demonstrating the on-device generation within seconds~\cite{castells2024edgefusion,li2024snapfusion,zhao2023mobilediffusion,sui2024bitsfusion}. 
However, these models are constrained to low-resolution output, \ie, $512^2$ px. To enable efficient high-resolution generation, SANA~\cite{sana} and LinFusion~\cite{liu2024linfusion} incorporate linear attention~\cite{katharopoulos2020transformers,cai2023efficientvit,dao2024transformers} to achieve 1K  generation on laptop GPUs. In contrast, we target a broader range of platforms, supporting high-resolution generation (\eg, 1K) directly on mobile devices.

\noindent\textbf{Knowledge Distillation in Diffusion Models.}
In the context of diffusion models, previous works focus on distilling large, high-capacity teacher models into more compact, efficient student models within a \emph{homogeneous} architecture~\cite{kim2023bk,liu2024linfusion}.
They reduce the complexity of the model by removing certain components like attention~\cite{vaswani2017attention} or residual blocks~\cite{he2016deep}, while maintaining the architectural structure. However, our approach diverges from this trend by utilizing a \emph{heterogeneous} architecture for more aggressively efficient yet challenging distillation.

\noindent\textbf{Adversarial Step Distillation} uses the techniques from adversarial training~\cite{goodfellow2014generative} to reduce the number of diffusion steps, while maintaining high image quality~\cite{sauer2024adversarial,sauer2024fast}.
For example, UFOGen~\cite{xu2024ufogen} employs a diffusion-GAN formulation~\cite{wang2022diffusion, xiao2021tackling, xu2023semi} to significantly reduce inference time while maintaining competitive performance. DMD2~\cite{yin2024improved} builds on prior distillation methods by using distribution matching with adversarial loss. Different from exiting works, we conduct step-distillation on a very compact model and train the model along with the knowledge distillation.

\begin{figure*}[!ht]
\begin{center}
\includegraphics[width=1.\textwidth]{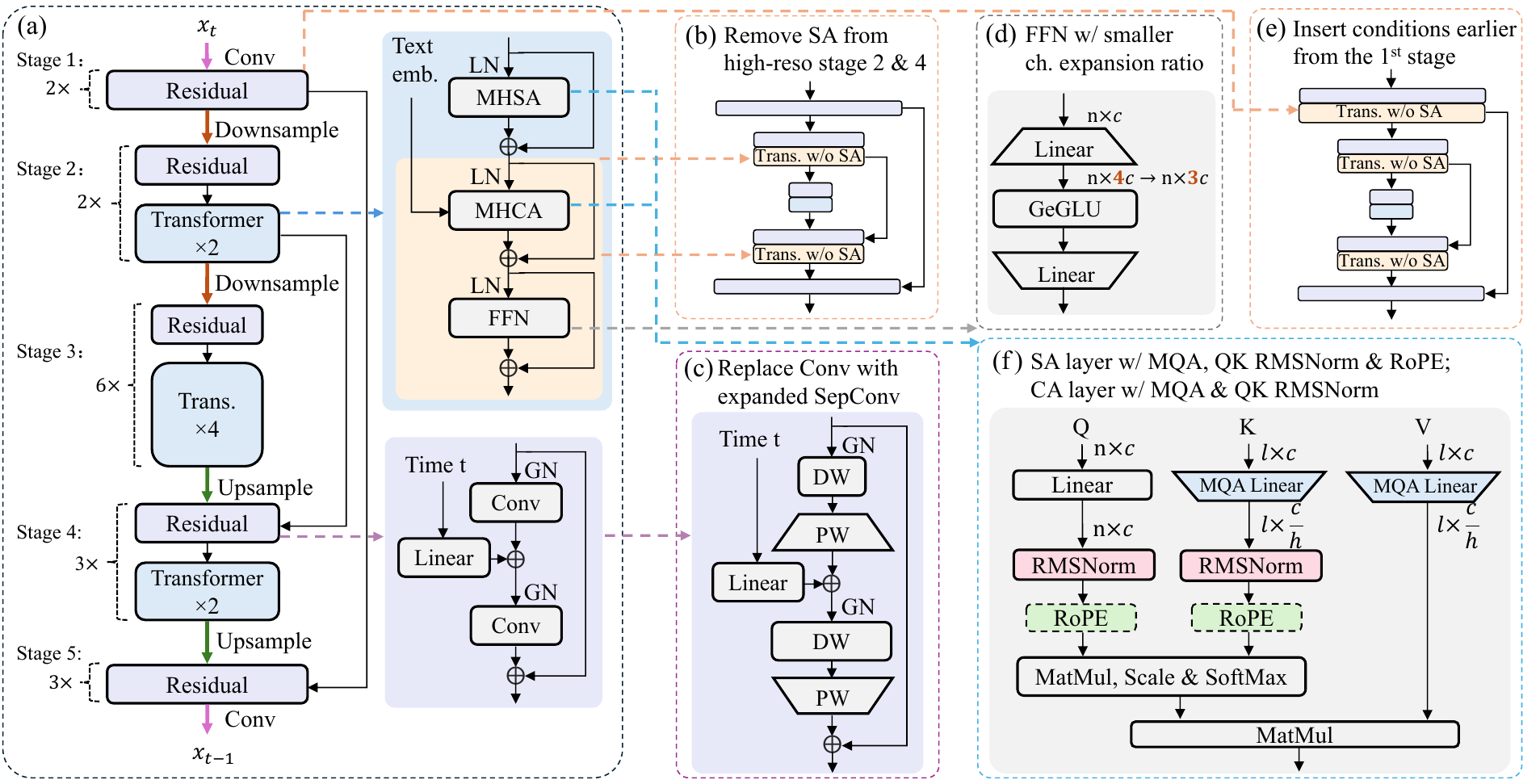}
\end{center}
\vspace{-0.5cm}
\caption{\textbf{Efficient UNet.} Starting from a thinner and shorter version of the UNet from SDXL (as in (a)), we explore a series of architectural changes, \ie, (b)--(f), to develop a smaller and faster model while retaining high-quality generation performance, as evaluated in \cref{fig:unet_ablation}.  
}\vspace{-0.4cm}
\label{fig:unet_architecture}
\end{figure*}

\begin{figure*}[ht]
\begin{center}
\includegraphics[width=.95\textwidth]{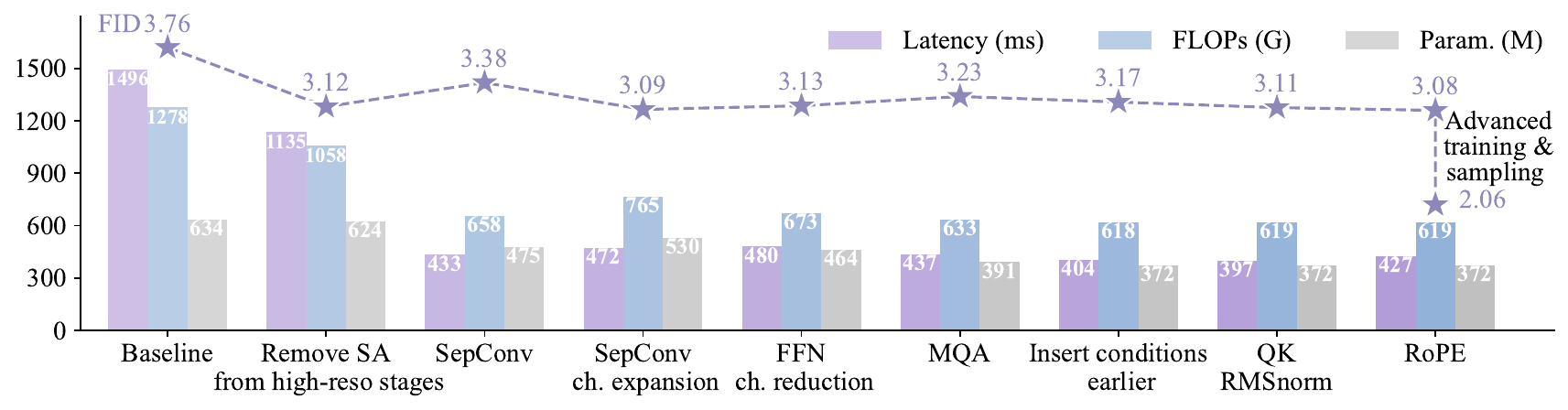}
\end{center}
\vspace{-0.6cm}
\caption{\textbf{Comparisons of Performance and Efficiency for various Design Choices of Efficient UNet.} The generation quality is evaluated using FID calculated on ImageNet-1K for $256^2$ px generation. The efficiency metrics include model parameters, latency, and FLOPs. FLOPs and latency (on iPhone 15 Pro) are measured with a $128\times128$ latent, equivalent to a $1024\times1024$ decoded image, for one forward pass.
We show the architecture enhancements that improve any of the metrics without hurting others. 
}\vspace{-0.3cm}
\label{fig:unet_ablation}
\end{figure*}

%% file: sec/3_method.tex
\section{Method}
In this section, we present how to craft and train a highly efficient T2I model for high-resolution generation. Specifically, starting from the architecture designed in the latent diffusion model~\cite{rombach2022high}, we optimize both denoising backbone (\cref{sec:efficienunet}, \cref{fig:unet_architecture}, and \cref{fig:unet_ablation}) and autoencoder (\cref{sec:efficient_decoder} and \cref{fig:ae_decoder_architecture}) to make them compact and fast, even on mobile devices. We then propose the improved training recipe and knowledge distillation (\cref{sec:training} and \cref{fig:knowledge_distillation}), empowering a high-performance T2I model. Lastly, we introduce our step distillation to significantly reduce the number of denoising steps for a faster T2I model (\cref{sec:step_distillation} and \cref{fig:step_distillation}).

\subsection{Efficient UNet Architecture}\label{sec:efficienunet}
Here we describe the design choices for the denoising UNet.

\noindent\textbf{Baseline Architecture.}
We choose UNet from SDXL~\cite{podell2023sdxl} as the baseline (Fig.~\ref{fig:unet_architecture}(a)) for our backbone since it has superior efficiency and faster convergence~\cite{li2024scalability} than pure transformer-based models~\cite{chen2023pixart,chen2024pixart}. We adjust the UNet into a thinner and shorter model (\ie, reducing the number of transformer blocks from $[0, 2, 10]$ in three stages to $[0, 2, 4]$ and their channel dimensions from $[320, 640, 1280]$ to $[256, 512, 896]$), and iterate design choices on top of it.

\noindent\textbf{Evaluation Metrics.} We train models on ImageNet-1K~\cite{deng2009imagenet} under class-conditional generation for $120$ epochs, unless specified otherwise, and report the FID score~\cite{fid} for $256^2$ px generation. Similarly to existing work~\cite{kag2024ascan}, 
we inject the class conditions through a text template ``a photo of \textless class name\textgreater''. 
We then encode it with a light text encoder to align the pipeline for the T2I generation. 
We also calculate the number of parameters for different models, floating point operations (FLOPs) (measured on a latent size of $128\times128$, equivalent to a $1024\times1024$ image after decoding), and the runtime on mobile device (tested on iPhone 15 Pro). Detailed training and evaluation settings can be found in Supplemental Materials. In the following, we introduce the key architectural changes that improve the model.

\noindent\textbf{Remove Self-Attention from High-Resolution Stages.} Self-attention (SA) layer is restricted by its quadratic computational complexity, incurring a heavy computational cost and high memory consumption for high-resolution input. As such, we keep the SA layer only in the lowest-resolution stage while removing it from other higher-resolution stages, \ie, Fig.~\ref{fig:unet_architecture} (b). This leads to $17\%$ fewer FLOPs and $24\%$ latency reduction, as shown in Fig.~\ref{fig:unet_ablation}. Interestingly, we even observe a performance improvement, \ie, FID decreased to $3.12$ from $3.76$. We hypothesize that models with SA in high-resolution stages converge more slowly.

\noindent\textbf{Replace Conv with Expanded Separable Conv.} Regular convolution (Conv) is redundant in both parameters and computation. To address this, we replace all Conv layers with the separable convolution (SepConv)~\citep{howard2017mobilenets}, composed of a depthwise convolution (DW) followed by a pointwise convolution (PW), as shown in Fig.~\ref{fig:unet_architecture} (c). This replacement reduces parameters by $24\%$ and latency by $62\%$, but also leads to a performance drop (FID increases from $3.12$ to $3.38$). To address the issue, we expand the intermediate channels. Specifically, the number of channels after the first PW layer is increased with an expansion ratio, and reduced back to the original number after the second PW layer. The expansion ratio is set to $2$ to balance the trade-off between performance, latency, and model parameters. Such a design aligns our residual block with the recently proposed Universal Inverted Bottleneck (UIB) block~\cite {qin2024mobilenetv4}. As a result, our model achieves $15\%$ fewer parameters, $27\%$ less computation, and $2.4\times$ speedup, while obtaining a lower FID.

\noindent\textbf{Trim FFN Layers.} For the layers in the feed-forward network (FFN), the hidden channel expansion ratio is set to $4$ by default and further doubled by using the gated unit. This substantially inflates the model parameters, computation, and memory usage. Following MobileDiffusion~\citep{zhao2023mobilediffusion}, we examine the efficacy of simply reducing the expansion ratio, as shown in ~\cref{fig:unet_architecture} (d). We show that reducing the expansion ratio to $3$ preserves comparable FID performance while reducing both parameters and FLOPs by $12\%$.

\noindent\textbf{Replace MHSA with MQA.} Multi-Head Self-Attention (MHSA) requires multiple sets of keys and values for each attention head. In contrast, Multi-Query Attention (MQA)~\citep{shazeer2019fast} is more efficient by sharing a single set of keys and values across all heads.
Replacing MHSA with MQA reduces parameters by $16\%$ and latency by $9\%$, with minimal impact on performance. Interestingly, the $9\%$ saving in latency exceeds the $6\%$ decrease in FLOPs, as the reduced memory access enables higher computational intensity.

\noindent\textbf{Inject Conditions to the First Stage.} 
Cross-attention (CA) blends the conditional information (\eg, textural description) along with the spatial features to generate images that align with the condition. However, the UNet of SDXL only applies CA in transformer blocks starting from the second stage, resulting in the missing conditional guidance for the first stage.
In response, we propose to introduce the conditional embeddings from the very first stage, as in Fig.~\ref{fig:unet_architecture}(e). Specifically, we replace the residual blocks with transformer blocks that include CA and FFN while without SA layers. This adjustment makes the model smaller, faster, and more efficient while improving FID.

\noindent\textbf{Employ QK RMSNorm and RoPE Positional Embeddings.} We extend two advanced techniques developed originally for language models, Query-Key (QK) Normalization~\citep{henry2020query} with RMSNorm~\citep{zhang2019root} and Rotary Position Embedding (RoPE)~\citep{su2024roformer}, to enhance the model (Fig.~\ref{fig:unet_architecture} (f)). RMSNorm, applied after the Query-Key projection in the attention mechanism, reduces the risk of softmax saturation without sacrificing model expressiveness while stabilizing training for faster convergence. In addition, we adapt RoPE from one dimension to two dimensions for better supporting higher resolution since it significantly mitigates artifacts like repeated objects.
Together, RMSNorm and RoPE introduce negligible computational and parameter overhead, while offering measurable gains in FID performance.

\noindent\textbf{Discussion.} The above optimization results in an efficient and powerful diffusion backbone capable of generating high-resolution images on mobile devices. 
Before proceeding with large-scale T2I training, we compare the capacity of our model against existing works on ImageNet-1K. We train the model for $1,000$ epochs by following the setting from prior work~\cite{rombach2022high}.
We evaluate the model using varied CFG~\cite{cfg,kynkaanniemi2024applying} across different inference timesteps. As shown in ~\cref{tab:imagenet_comparison}, our efficient UNet achieves comparable FID to SiT-XL~\citep{ma2024sit}, while being almost $45\%$ smaller.

\begin{table}[ht]
\centering
\footnotesize
\vspace{-0.1in}
\caption{ \textbf{Class-conditional image generation on ImageNet $256\times256$} with CFG. FLOPs are calculated for one forward pass.}
\label{tab:imagenet_comparison}
\vspace{-0.1in}
\begin{tabular}{lccc}
\toprule
Model       & Param (M) & FLOPs (G) & FID$\downarrow$ \\ \midrule
LDM-4 \citep{rombach2022high}      & 400 & 104  & 3.60     \\
UViT-L \citep{bao2023all}      & 287  & 77  & 3.40   \\
UViT-H  \citep{bao2023all}     & 501  & 133  & 2.29   \\
DiT-XL \citep{peebles2023scalable}      & 675  & 119  & 2.27    \\
SiT-XL  \citep{ma2024sit}     & 675  & 119  & 2.06     \\
\midrule
Ours        & 372  & 38 & 2.06  \\ 
\bottomrule
\end{tabular}
\vspace{-0.15in}
\end{table}

\subsection{Tiny and Fast  Decoder}\label{sec:efficient_decoder}
  
\begin{figure}[ht]
\begin{center}
\includegraphics[width=1\linewidth]{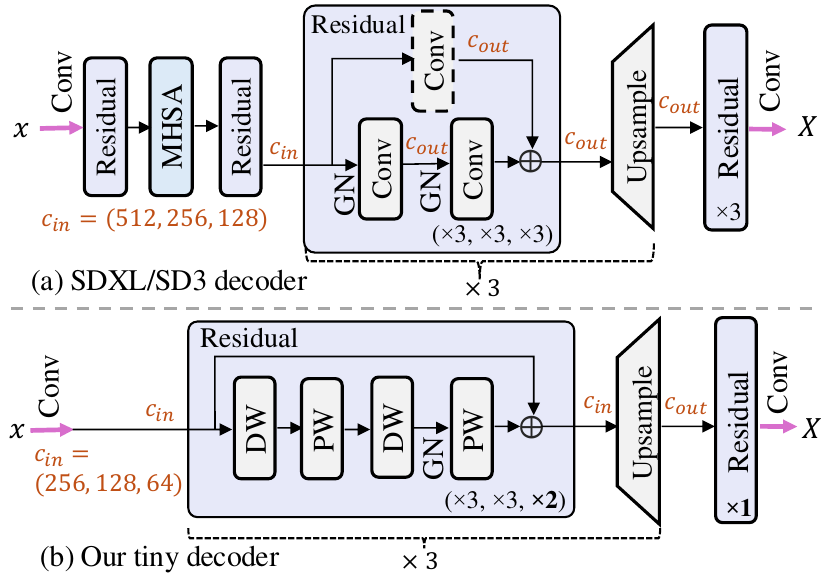}
\end{center}
\vspace{-0.5cm}
\caption{\textbf{Comparisons of Decoder Architecture} between (a) SDXL/SD3 decoder and (b) our tiny decoder.}
\label{fig:ae_decoder_architecture}
\vspace{-0.1in}
\end{figure}

Besides the denoising model, the decoder also takes a significant ratio of total runtime, especially for on-device deployment~\cite{li2024snapfusion,zhao2023mobilediffusion}. Here we introduce a new architecture of decoder (\cref{fig:ae_decoder_architecture}) for efficient high-resolution generation.

\noindent\textbf{Baseline Decoder.} We use the autoencoder (AE) from SD3~\cite{esser2024scaling} as our baseline model (\ie, the same encoder from SD3 AE), due to its superior reconstruction quality. The AE maps an image $X\in \R^{H\times W\times 3}$ into a lower-dimensional latent $x \in \R^{\frac{H}{f} \times \frac{W}{f} \times c}$ ($f, c$ as $8,16$ in SD3). The encoded latent $x$ is then decoded back to an image through a decoder. 
For high-resolution generation, we observe that the decoder in SD3 is very slow on mobile devices. Specifically, it encounters out-of-memory (OOM) errors when generating a $1024^2$ px image on both the ANE processor of the iPhone 15 Pro and the mobile GPU (\cref{tab:vae_size_latency}).
To overcome the latency issue, we propose a much smaller and faster decoder.

\begin{table}[ht]
\caption{\textbf{Performance Comparison of Decoder.} PSNR is calculated on COCO 2017 validation set~\cite{lin2014microsoft}. FLOPs and latency (on iPhone 15 Pro) are measured for decoding a $128\times128$ latent into a $1024\times1024$ image. The decoder from SDXL and SD3 fail to run on the neural engine of mobile, resulting in a huge runtime.
}
\label{tab:vae_size_latency}
\centering
\vspace{-0.1cm}
\resizebox{1\linewidth}{!}{
\begin{tabular}{lcccccc}
 \toprule
Decoder & Ch & PSNR & Param (M) & FLOPs (G) & \begin{tabular}[c]{@{}l@{}}Latency (ms)\\ on ANE \end{tabular}  & \begin{tabular}[c]{@{}l@{}}Latency (ms)\\ on GPU \end{tabular}   \\
\midrule
SDXL~\citep{podell2023sdxl}    & 4  & 24.89 & 49.49    & 4970  & OOM &  9469     \\ 
SD3~\citep{esser2024scaling}    &  16  & 27.92 & 49.55  & 4970  & OOM &  OOM    \\ 
\hline
Ours &  16 &  27.85 & 1.38   &  224 & 174    &   -    \\ 
\bottomrule
\end{tabular}}
\vspace{-0.3cm}
\end{table}

\noindent\textbf{Efficient Decoder.}
We conduct a series of experiments to decide the efficient decoder with the following key changes compared with the baseline architecture:
\begin{enumerate}
    \item We remove attention layers to greatly reduce peak memory without a noticeable impact on decoding quality.
    \item We keep a minimal amount of GroupNorm (GN) to find a trade-off between latency and performance (\ie, mitigating the color shifting).
    \item We make the decoder thinner (\ie, fewer channels or narrower width) and replace Conv with SepConvs.
    \item We use fewer residual blocks in high-resolution stages.
    \item We remove the Conv shortcut in residual blocks and use the upsampling layer for channel transition.
\end{enumerate}
\noindent\textbf{Training of the Decoder.}
We train our decoder with the mean squared error (MSE) loss, lpips loss~\cite{zhang2018perceptual}, adversarial loss~\cite{goodfellow2014generative}, and discard the KL term~\cite{kingma2013auto} as the encoder is fixed.  The decoder is trained on $256^2$ image patches with a batch size of $256$ and for $1$M iterations.  As in \cref{tab:vae_size_latency}, our tiny decoder achieves a competitive PSNR score for reconstruction, while being
$35.9\times$ \emph{smaller} and {$54.4\times$} \emph{faster} for high-resolution generation on mobile devices compared to conventional ones (\eg, the decoder from SDXL and SD3).

\noindent\textbf{Discussion of Total On-Device Latency.} We finally measure T2I model latency for a $1024^2$ px generation on iPhone 16 Pro-Max. The decoder takes $119$ms, and the per-step latency for the UNet is $274$ms. This results in a $1.2\sim2.3$s runtime for $4$ to $8$ step generation. Note that text encoder runtime is negligible compared to other components~\cite{li2024snapfusion}.

\begin{figure*}[t]
    \begin{minipage}{0.32\textwidth}
    \includegraphics[width=\textwidth]{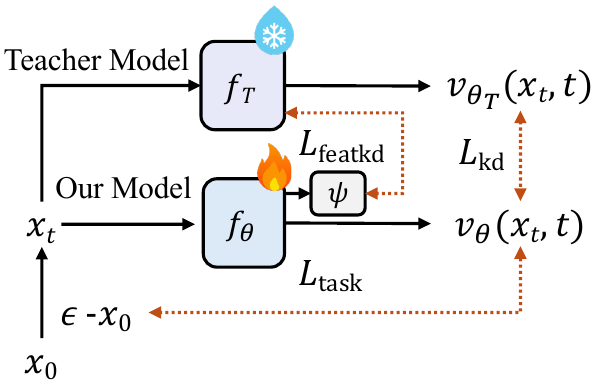}
    \caption{Overview of Multi-level Knowledge Distillation, where we perform output distillation and feature distillation.} \label{fig:knowledge_distillation}
    \end{minipage}\vspace{-0.05in}
    \hspace{2mm}
    \begin{minipage}{0.32\textwidth}
    \includegraphics[width=\textwidth]{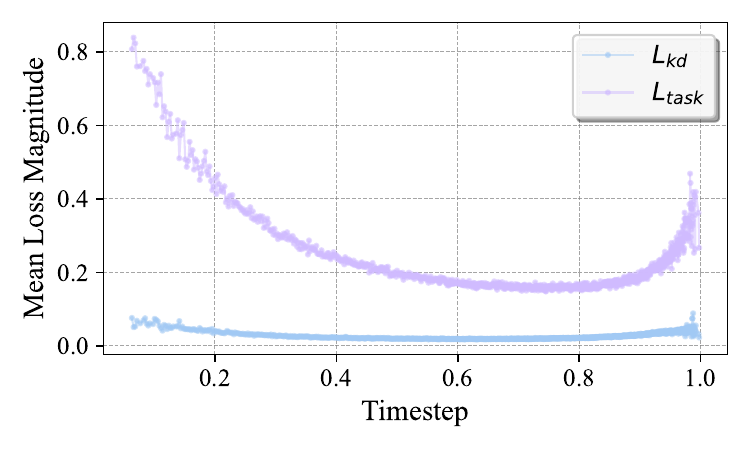}
    \caption{Mean loss magnitude for task loss $\mathcal{L}_{\mathrm{task}}$ and output distillation loss $\mathcal{L}_{\mathrm{kd}}$.} \label{fig:loss-scale}
    \end{minipage}\vspace{-0.05in}
    \hspace{2mm}
    \begin{minipage}{0.32\textwidth}
    \includegraphics[width=\textwidth]{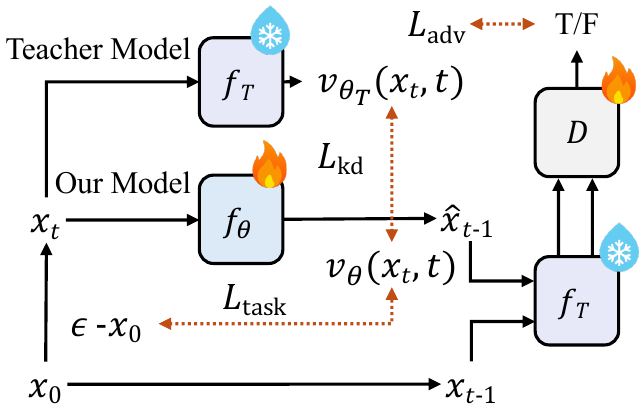} 
    \caption{Overview of Adversarial Step Distillation. Output distillation and distribution-matching distillation are performed.} \label{fig:step_distillation}
    \end{minipage}\vspace{-0.05in}
\end{figure*}%

\subsection{Training Recipe and Multi-Level Distillation}\label{sec:training}

To improve the generation quality of our efficient diffusion model, we propose a series of training techniques.

\noindent\textbf{Flow-based Training and Inference.}
Rectified Flows (RFs)~\cite{liu2022flow, lipman2022flow} define the forward process as straight paths connecting the data distribution to a standard normal distribution, \ie, 
\begin{equation} \label{eq:forward}\small
x_t = (1 - \sigma_t) x_0 + \sigma_t \epsilon,
\end{equation}
where  $x_0$  is the clean (latent) image, $t$ is the timestep, $\sigma_t$ is a timestep-dependent factor, and $\epsilon$ is random noise sampled from $\mathcal{N}(0, I)$. The denoising UNet is formulated to predict a velocity field with the objective as
\begin{equation} \label{eq:loss_task}\small
\mathcal{L}_{\mathrm{task}} = \mathbb{E}_{\epsilon \sim \mathcal{N}(0, I), t} \Big[ ||(\epsilon - x_0)-v_{\theta}(x_t, t)||_2^2 \Big],
\end{equation}
where $v_\theta(x_t, t)$ is the predicted velocity from UNet parameterized by $\theta$. To further enhance training stability, we apply logit-normal sampling~\cite{esser2024scaling} for the timestep during training, which assigns more samples to the intermediate steps.
In the inference stage, we use the Flow-Euler sampler~\cite{euler1768institutionum}, which predicts the next sample based on the velocity, \ie,
\begin{equation}\small
x_{t-1} = x_t + (\sigma_{t-1} - \sigma_t) \cdot v_\theta(x_t, t).
\end{equation}
To achieve a lower signal-to-noise ratio on high-resolution (\ie, $1024^2$ px) images, we apply a timestep shift similar to SD3~\cite{esser2024scaling} to adjust the scheduling factor  $\sigma_t$  during both training and inference.

\noindent\textbf{Multi-Level Knowledge Distillation.} 
To improve the generation quality for compact models, one common practice for previous works is applying knowledge distillation to mimic the prediction of scaled-up teacher models~\cite{kim2023bk}. Benefiting from the aligned flow matching objective and (AE) latent space, powerful SD3.5-Large model~\cite{sd3.5-large} can be used as the teacher for output distillation. However, we still face challenges due to 1) the heterogeneous architecture between the U-Net and DiT, 2) the scale difference between the distillation loss and task loss, and 3) varying prediction difficulty across different timesteps. To tackle these, we propose a novel multi-level distillation loss and apply timestep-aware scaling to stabilize and accelerate the convergence of distillation. 
The overview of our knowledge-distillation scheme is shown in \cref{fig:knowledge_distillation} and detailed techniques are elaborated as follows.

Aside from the task loss defined in Eq.~\ref{eq:loss_task}, the major objective for knowledge distillation is to supervise our model $\theta$ directly with the output of teacher model $\theta_T$, which can be indicated as
\begin{equation} \label{eq:loss_out_kd}
\small
\mathcal{L}_{\mathrm{kd}} = \mathbb{E}\Big[ ||v_{\theta_T}(x_t, t)-v_{\theta}(x_t, t)||_2^2 \Big].
\end{equation}
Given the capacity gap between the teacher and our model, applying output-level supervision alone leads to instability and slow convergence. Therefore, we further incorporate a cross-architecture feature-level distillation loss as
\begin{equation} \label{eq:loss_feat-kd}
\small
\mathcal{L}_{\mathrm{featkd}} = \mathbb{E}\Big[ \sum_{(l_T, l)}||f_{\theta_T}^{l_T}(x_t, t)-\psi(f_{\theta}^{l}(x_t, t))||_2^2 \Big],
\end{equation}
where $f_{\theta_T}^{l_T}(\cdot)$ and $f^l(\cdot)$ indicate the feature output from the $l_T$-th layer and $l$-th layer in teacher model and student model, respectively. Different from previous work~\cite{liu2024linfusion,kim2023bk}, we consider cross-architecture distillation from a DiT to UNet. Since the richest information in transformers sits around the last layer, we set the distillation target to this layer in both models, and use a lightweight trainable projector $\psi(\cdot)$ with only two Conv layers to map the student feature to match the dimension of teacher feature. The proposed feature-level distillation loss provides additional supervision to the student model, leading to faster alignment to the generation quality of the teacher model.

\noindent\textbf{Timestep-Aware Scaling.} Weighting multiple objectives has been a major challenge in knowledge distillation, especially in diffusion models. The overall training objectives from previous works~\cite{sui2024bitsfusion,liu2024linfusion,kim2023bk} are simple linear combination of multiple loss term, \ie,
\begin{equation} \label{eq:loss_final}
\small
\mathcal{L} = \mathcal{L}_{\mathrm{task}} + \lambda_1\mathcal{L}_{\mathrm{kd}} + \lambda_2 \mathcal{L}_{\mathrm{featkd}},
\end{equation}
where the weighting coefficient $\lambda_1$ and $\lambda_2$ are empirically set to constant. However, this baseline setting fails to consider the \textit{prediction difficulty} in various time steps. We investigate the distribution of empirical risk magnitude of $\mathcal{L}_{\mathrm{task}}$ and $\mathcal{L}_{\mathrm{kd}}$ across different timestep $t$ during model training. \cref{fig:loss-scale} illustrates that, in intermediate steps, \textit{prediction difficulty} are lower compared to $t$ closer to $0$ or $1$. 

Building on this important observation, we propose a timestep-aware scaling of the objective to close the gap in loss magnitude across different values of $t$ and to account for prediction difficulties at each timestep, as follows:
\begin{equation} \label{eq:scaling} \small
\mathcal{S}(\mathcal{L}_{\mathrm{task}}, \mathcal{L}_{\mathrm{kd}}) = \mathbb{E}_t \Big[\lambda(t) \cdot \mathcal{L}^t_{\mathrm{task}} + \big(1-\lambda(t)\big) \frac{|\mathcal{L}^t_{\mathrm{task}}|}{|\mathcal{L}^t_{\mathrm{kd}}\\|} \cdot \mathcal{L}^t_{\mathrm{kd}} \Big],
\end{equation}
where $\lambda(t)$ is the normalized standard (location $0$, scale $1$) logit-norm density function and $|\cdot|$ indicates the magnitude. In $\mathcal{S}$, we first ensure the same scale between task loss and distillation loss across different $t$, then apply more teacher supervision where \textit{prediction difficulty} is higher (\ie, $t$ closer to $0$ or $1$), and more real data supervision where \textit{prediction difficulty} is lower (\ie, intermediate timesteps). The proposed scheme considers the variation of timestep $t$ and helps accelerate the distillation training. The final multi-level distillation objective $\mathcal{L}_{\mathrm{MD}}$ is defined as
\begin{equation} \label{eq:loss_final_ours}
\small
\mathcal{L}_{\mathrm{MD}} = \mathcal{S}(\mathcal{L}_{\mathrm{task}}, \mathcal{L}_{\mathrm{kd}}) +\mathcal{S}(\mathcal{L}_{\mathrm{task}}, \mathcal{L}_{\mathrm{featkd}}).
\end{equation}

\subsection{Step Distillation}\label{sec:step_distillation}

We take one step further to enhance the sampling efficiency of our model following a distribution-matching-based step distillation scheme. Following  Latent Adversarial Diffusion Distillation (LADD)~\cite{sauer2024fast}, we use a diffusion-GAN hybrid structure to distill our model into fewer steps with the optimization objective as
\begin{equation} \label{eq:stepkd_obj}
\small
\begin{aligned}
    &\min_{D_{\theta_T}}\max_{G_\theta}\mathbb{E} \Bigl[  [\log(D_{\theta_T}(x_{t-1}, t))] \\
    &+ [\log(1-D_{\theta_T}(x^\prime_{t-1}, t))] - \mathcal{S}(\mathcal{L}_{\mathrm{task}}, \mathcal{L}_{\mathrm{kd}}) \Bigl],
\end{aligned}
\end{equation}
where $D_{\theta_T}$ is the discriminator model partially initialized with pretrained fewer-step teacher model $\theta_T$ (SD3.5-Large-Turbo~\cite{sd3.5-large-turbo}). The large-scale teacher model are only used as the feature extractor and are frozen during distillation. We only train a few linear layers in the discriminator after feature extraction. We sample $x_{t-1}\sim q(x_{t-1}|x_0)$ and $x_{t-1}^\prime \sim q(x_{t-1}|x_0^\prime)$, where $x_0^\prime$ is the prediction of our denoising generator\footnote{For simplicity, we use the $x_0$ prediction in our derivation, and $v$ prediction of rectified flow-based training would not break the formulation.} $G_\theta(x_t,t)$ as our student model, and $q(x)$ is the forward process of diffusion model defined in Eq.~\ref{eq:forward}. The objective consists of an adversarial loss to match noisy samples at time step $t-1$ and the output-level distillation loss $\mathcal{S}(\mathcal{L}_{\mathrm{task}}, \mathcal{L}_{\mathrm{kd}})$ after applying timestep-aware scaling. The proposed step distillation, 
visualized in Fig.~\ref{fig:step_distillation}, can be interpreted as training a diffusion model with adversarial refinement and knowledge distillation, where teacher guidance serves as an additional inductive bias. This advanced step distillation empowers our compact model for high-quality generation, with only a few denoising steps.

%% file: sec/4_experiments_v2.tex
\section{Experiments}
\noindent\textbf{Model Details.} Our T2I pipeline consists of the efficient UNet (Sec.~\ref{sec:efficienunet}) and the efficient encoder-decoder model (Sec.~\ref{sec:efficient_decoder}). To obtain text embeddings from the input prompt, we leverage multiple text encoders, namely light-weight CLIP-L~\cite{radford2021learning}, CLIP-G~\cite{radford2021learning}, and the large Gemma-2-2b language model~\cite{team2024gemma}. We follow SD3~\cite{esser2024scaling} strategy to combine these three text-encoders into a unified rich textual embedding. To enable classifier-free guidance~\cite{ho2022classifier}, we employ these embeddings with an individual drop-out probability such that we can use an arbitrary subset of the encoders during inference. This allows us to deploy one or more encoders based on the resource constraints. 

\noindent\textbf{Training Recipe.} Similar as prior work~\cite{kag2024ascan}, we use a multi-stage strategy to train our UNet model from scratch. First, we pre-train the model using the ImageNet-1K~\cite{deng2009imagenet} at $256$ resolution as described in Sec.~\ref{sec:efficienunet}. 
Second, we fine-tune this model
in a progressive manner from $256 \to 512 \to 1024$ resolutions. 
Third, we employ knowledge distillation with our timestep-aware scaling (Sec.~\ref{sec:training}) to improve the finer details in our models using a much larger teacher model (SD3.5-Large~\cite{sd3.5-large}) and all three text-encoders. 
Finally, we obtain a few-step model through step distillation using SD3.5-Large-Turbo~\cite{sd3.5-large-turbo} model as the teacher. We optimize the rectified-flow~\cite{esser2024scaling} objective using AdamW optimizer~\cite{diederik2014adam} to train our UNet backbone. 

\noindent\textbf{Hyper-parameters.} We sample the timesteps in the flow-matching using the logit-normal distribution with $(0,1)$ as the location and scale parameters. We use a time shift of $3$ for both training and inference. For unconditional diffusion guidance~\cite{ho2022classifier}, we set the outputs of each of the three text encoders independently to zero with a probability of $46.4$\%, such that we roughly train an unconditional model in $10$\% of all steps. See the Supplemental Materials for details.

\subsection{Evaluation}
\noindent\textbf{Quantitative Benchmarks.} We use GenEval~\cite{ghosh2024geneval} and DPG-Bench~\cite{hu2024ella} benchmarks to evaluate the text-to-image alignment of our model on short and long prompts, respectively. We report CLIP score on a 6K subset of MS-COCO validation data~\cite{lin2014microsoft}. In addition, to measure the aesthetic quality of our model, we compute the Image Reward~\cite{xu2024imagereward} score on selected PixArt prompts~\cite{chen2023pixart}. Tab.~\ref{tab:quantitative_benchmarks} lists our performance alongside existing state-of-the-art T2I baselines. We provide additional details in Supplemental Materials. We highlight the salient observations below:
\begin{itemize}
\item Our $0.38$B parameter model achieves even better performance than significantly larger models such as SDXL ($2.6$B), Playground ($2.6$B), and IF-XL ($5.5$B).
\item KD non-trivially improves the prompt following ability of the base model as illustrated by an absolute five-point increase in DPG-Bench and GenEval scores.
\item In terms of aesthetic performance, our model has similar Image Reward scores as Playground models~\cite{playground-v2, li2024playground}.
\end{itemize}

\begin{table}[ht]
\caption{\textbf{Evaluation on Quantitative Benchmarks.} We list the scores on GenEval, DPG-Bench, CLIP score on COCO, and Image Reward on aesthetic prompts. We report the parameters for the UNet/DiT backbone in the Param column. Throughput (samples/s) is measured on a single 80GB A100 GPU using the largest batch size supported for each model in a practical scenario to generate $1024^2$ px images.
Here our sampling step is set to be 28.
}
\label{tab:quantitative_benchmarks}
\centering
\vspace{-0.1in}
\resizebox{\linewidth}{!}{
\begin{tabular}{l|cc|cccc}
\toprule
Model       & Param  & \begin{tabular}[c]{@{}c@{}}Throughput \end{tabular} & GenEval $\uparrow$ & DPG $\uparrow$ & CLIP $\uparrow$ & \begin{tabular}[c]{@{}c@{}}Image \\ Reward $\uparrow$\end{tabular} \\ \midrule
PixArt-$\alpha$~\cite{chen2023pixart}   & 0.6B  & 0.42    & 0.48 & 71.1   & 0.316  & 1.15  \\
PixArt-$\Sigma$~\cite{chen2024pixart}   & 0.6B  & 0.46    & 0.53 & 80.5   & 0.317    & 1.13   \\
SD-1.5~\cite{sd1.5}                     & 0.9B  & -    & 0.43 & 63.2   & 0.287  & 0.19  \\
SD-2.1~\cite{sd2.1}                     & 0.9B  & -    & 0.50 & 64.2   & 0.281  & 0.29  \\
Sana~\citep{sana}                 & 1.6B  & 1.00    & \textbf{0.66} & \textbf{84.8}   & \underline{0.327}   & 1.25  \\
LUMINA-Next~\cite{zhuo2024lumina}       & 2.0B  & 0.06    & 0.46 & 74.6   & 0.309  & 0.88  \\
SDXL~\cite{podell2023sdxl}              & 2.6B  & 0.18    & 0.55 & 74.7   & 0.301  & 0.99  \\
Playgroundv2~\cite{playground-v2}    & 2.6B  & 0.18    &  0.59 & 74.5   & 0.317  & 1.25  \\
Playgroundv2.5~\cite{li2024playground}  & 2.6B  & 0.18    & 0.56 & 75.5   & 0.319  & \textbf{1.34}  \\
IF-XL~\citep{deepfloyd}                 & 5.5B  & 0.06    & \underline{0.61} & 75.6   & 0.311   & 0.65  \\
\midrule
Ours w/o KD                             & 0.38B  &  1.04   & \underline{0.61} & 76.3   & 0.321  &  1.20 \\ 
SnapGen (ours)                          & 0.38B &  1.04   & \textbf{0.66} & \underline{81.1}   & \textbf{0.332}  &  \underline{1.32} \\ 
\bottomrule
\end{tabular}}
\end{table}

\noindent\textbf{Qualitative Comparison.} To visually evaluate the image-text alignment and aesthetics, we compare the generated images from different T2I models in \cref{fig:teaser}. We observe that many existing models fail to fully capture the full prompt and miss important elements. Further, human generations often result in smoothened-out faces, leading to the loss of details. In contrast, our model generates much more photo-realistic images with better image-text alignment. 

\begin{figure*}[bht]
\begin{center}
\includegraphics[width=0.95\textwidth]{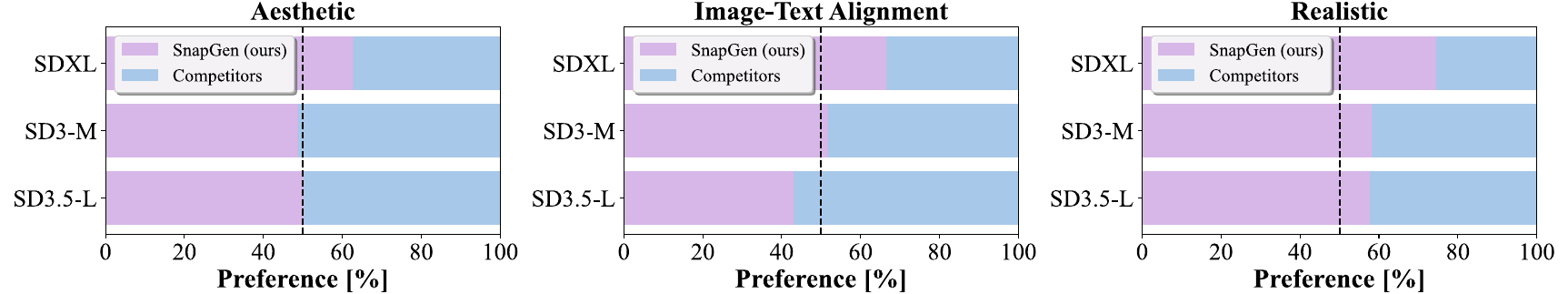}
\end{center}
\vspace{-0.2in}
\caption{\textbf{Human Evaluation.} We conduct a user study to compare images generated by our model against baselines on three attributes: aesthetic quality, text-image alignment, and realistic generations. Our model surpasses the quality of SDXL and SD3 models, while performing competitively against the teacher SD3.5-Large model.}
\label{fig:human_evaluation}\vspace{-0.1in}
\end{figure*}

\noindent\textbf{Human Evaluation.} For a thorough comparison between baselines, we perform a user study with the widely used Parti prompts~\cite{yuscaling}. We use SDXL, SD3-M, and SD3.5-Large models as the baselines and generate images using this prompt set. We ask the users to select images with better attributes between the baselines and our model. These attributes include image-text alignment, aesthetic quality, and realistic images. Fig.~\ref{fig:human_evaluation} shows that our model convincingly outperforms SDXL on all three attributes. Our model beats SD3 on image-text alignment and realistic generations, with a tie on aesthetic quality. Compared to the teacher (SD3.5-Large), we lag the teacher a bit behind on the image-text alignment, yet our model still has better realistic generations, and yields a toss-up on aesthetic quality. With this study, we can conclude our efficient T2I pipeline achieves generation quality that is quite comparable to the SD3.5-Large teacher that has $8.1$B parameters.

\begin{figure}[h]
    \centering
  \includegraphics[width=0.48\textwidth]{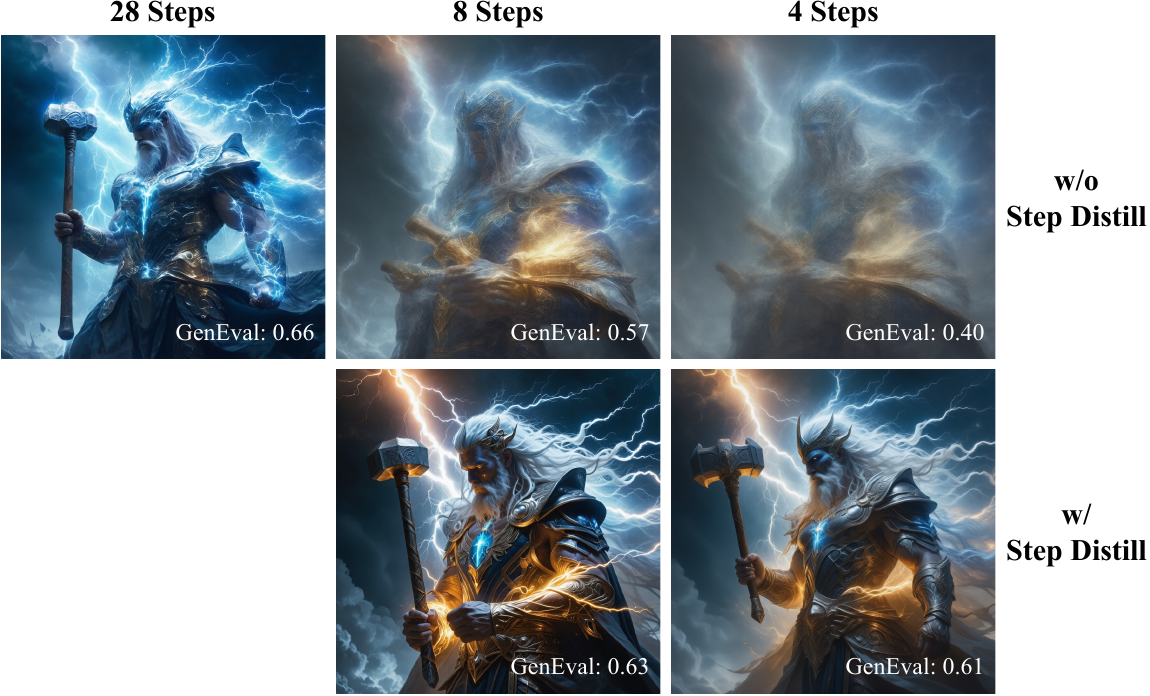}
    \vspace{-0.2in}
    \caption{Performance comparison of few-step generation for our model before (top) and after step distillation (bottom).} \label{fig:fewstep}\vspace{-0.05in}
\end{figure}

\noindent\textbf{Few-Step Generation.} After step-distillation (Sec.~\ref{sec:step_distillation}), our model can generate high-quality images within a few steps. 
Fig.~\ref{fig:fewstep} compares our model’s performance before and after step-distillation, with respective GenEval scores. The results demonstrate that our model, after step distillation, achieves comparable performance to the baseline model with 28 steps, even when using only 4 or 8 steps. While the few-step generation shows slight qualitative degradation compared to the 28-step baseline, it still outperforms most existing T2I models with significantly more inference steps, such as SDXL (50 steps) and PixArt-$\alpha$ (100 steps).

\vspace{-0.05in}
\vspace{-0.1in}

%% file: sec/5_conclusion.tex
\section{Conclusion}
In this work, we propose a novel and efficient T2I model for high-resolution generation on mobile phones. We systematically detail the process to obtain a tiny $379$M parameter UNet architecture along with an efficient latent decoder. We devise a novel training method consisting of multi-stage pre-training followed by knowledge distillation from a large teacher and adversarial step distillation. With these, we achieve an extremely efficient T2I model that comprehensively outperforms many existing multi-billion parameter models such as SDXL, Lumina-Next, and Playgroundv2. 

\noindent\textbf{Author Contribution Statement:}
\vspace{0.1in}
\begin{itemize}
    \item Dongting Hu designed and implemented the training pipeline, integrating various text encoders and incorporating a flow-matching objective to enable knowledge distillation from scalable DiT-based models (SD3.5). He prepared high-quality training data and trained the T2I base model, achieving an initial GenEval score of \emph{0.61}. He also built the distillation pipeline and contributed to multi-level knowledge distillation, significantly enhancing the model’s GenEval score to \emph{0.66} and improving generation quality based on human evaluations. His work on step distillation further enabled efficient high-quality generation, achieving a GenEval score of \emph{0.63} with 8-step generation and \emph{0.61} with 4-step generation. Additionally, he managed latent decoder training, ensuring close reconstruction quality to SD3 decoder, and facilitated on-device deployment. He developed the mobile app using the Core ML Diffusers framework, which achieved 1K resolution image generation on-device in approximately \emph{1.4} seconds, as demonstrated in the \href{https://snap-research.github.io/snapgen}{demo}.\\

    \item Jierun Chen developed the efficient UNet and AE decoder, enabling 1K resolution image generation on mobile devices \emph{for the first time}. On ImageNet class-conditional generation, the UNet achieves an FID score on par with the recent SiT-X model while reducing parameters by 45\% and compute resources by 68\%. The tiny decoder is $36\times$ \emph{smaller} and {$54\times$} \emph{faster} than conventional decoders (\eg, those from SDXL and SD3) for high-resolution mobile generation. He also initiated the early T2I diffusion training and contributed to the on-device deployment.\\

    \item Xijie Huang proposed the multi-level knowledge distillation scheme to improve the generation quality of our model, achieving comparable performance to the DiT-based teacher (SD3.5) across various quantitative benchmarks (\eg, boosting GenEval performance from 0.61 to \emph{0.66}) and human evaluation. He analyzed scale differences between distillation and task losses across timesteps, introducing a timestep-aware scaling operation. He also worked on adversarial step distillation to enable efficient and effective 4/8-step generation, leading to optimal latency on mobile devices.
\end{itemize}

%% file: sec/X_suppl.tex
\clearpage
\maketitlesupplementary
\setcounter{section}{0}
\setcounter{figure}{0}
\setcounter{table}{0}
\renewcommand{\thesection}{\Alph{section}}
\setcounter{section}{0}
\section{Demo on Mobile Devices}

\begin{figure}[h]
\begin{center}
    \captionsetup{type=figure}
    \includegraphics[width=1\linewidth]{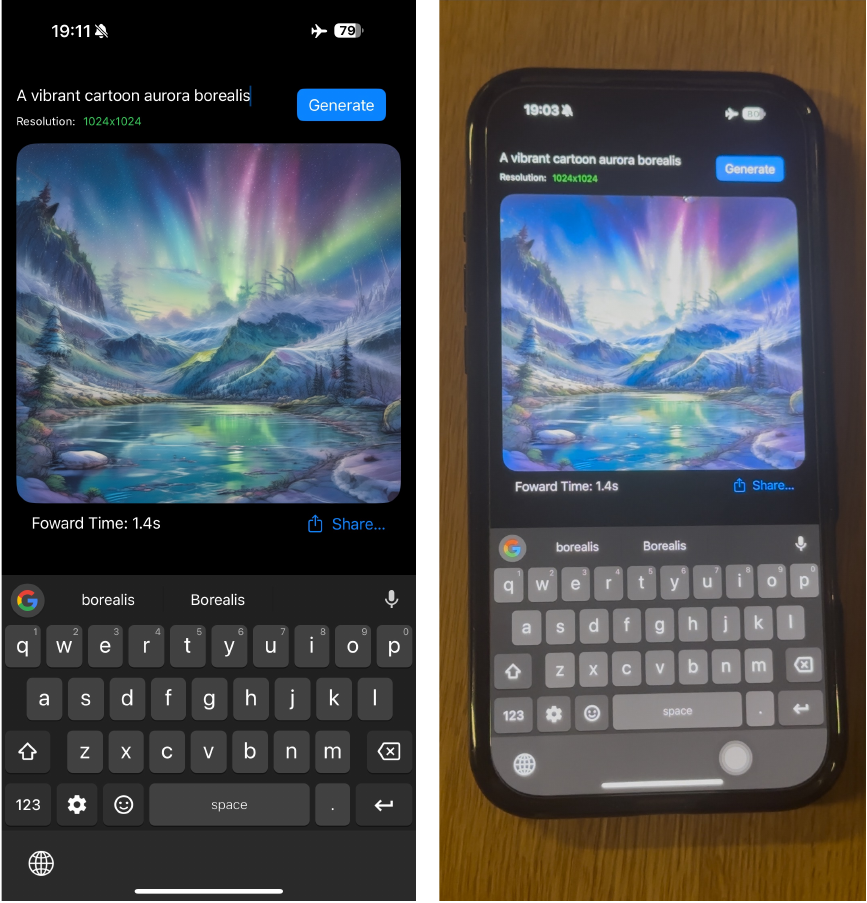}
        \caption{\textbf{Demo on iPhone 16 Pro-Max.} We report the forward time for each 4-step generation, excluding the model loading time.}
    \label{fig:demo}
\end{center}
\end{figure}

We present an on-device demo showcasing the capabilities of our efficient text-to-image model in generating high-resolution images (1024$\times$1024 pixels) directly on mobile phones. The application is implemented based on the open-source Swift Core ML Diffusers framework\footnote{https://github.com/huggingface/swift-coreml-diffusers}.
Upon launching the application, users can input textual prompts and generate corresponding images by clicking the ``Generate'' button. A screenshot of the deployed application is shown in~\cref{fig:demo}, and a more detailed demonstration can be found in the \href{https://snap-research.github.io/snapgen}{webpage}.

\section{Reconstruction by VAE Decoders}
\label{appendix:decoder_recon}

\begin{figure}[!t]
\begin{center}
\includegraphics[width=1\linewidth]{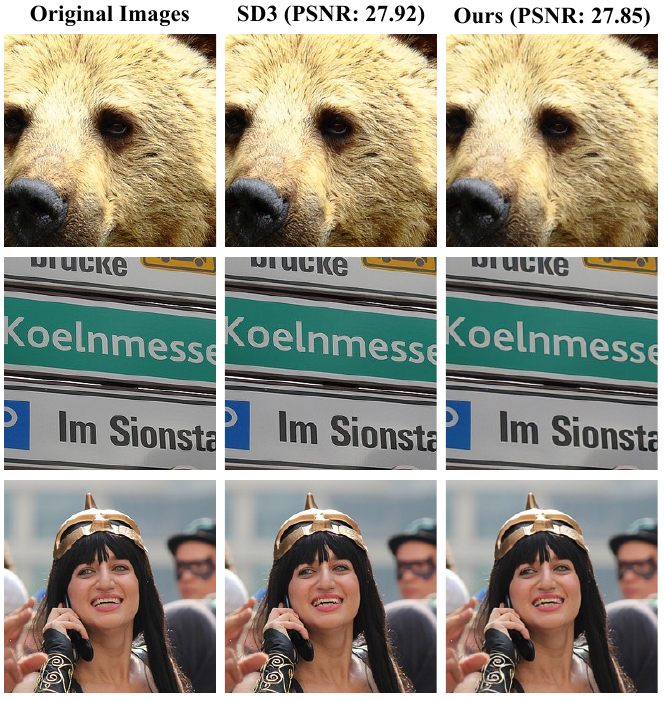}\vspace{-0.15in}
\end{center}
\caption{\textbf{Comparisons of Decoder Reconstruction} between SD3 decoder and our tiny decoder. Zoom in for better viewing.}\vspace{-0.1in}
\label{fig:vae_decoder}
\end{figure}

We compare reconstruction results between the SD3 VAE decoder (49.55M parameters) and our tiny decoder (1.38M parameters) in~\cref{fig:vae_decoder}. Despite being 36$\times$ smaller, our tiny decoder achieves competitively high visual quality across images with intricate textures, text, and human faces.

\begin{table*}[!t]
\caption{ \textbf{Detailed Results of GenEval Bench Comparisons.}}
\label{tab:geneval}
\centering
\resizebox{0.87\linewidth}{!}{
\begin{threeparttable}[b]
\begin{tabular}{lc|ccccccc}
\toprule
Model       & Param  & Overall $\uparrow$ &Single Object & Two Objects & Counting & Colors & Position & Color Attribution \\ \midrule
PixArt-$\alpha$~\cite{chen2023pixart}  & 0.6B  & 0.48    & 0.98  & 0.50 & 0.44     & 0.80   & 0.08     & 0.07 \\
PixArt-$\Sigma$~\cite{chen2024pixart}  & 0.6B  & 0.53    & 0.99  & 0.65 & 0.46     & 0.82   & 0.12     & 0.12 \\
SD-1.5~\cite{sd1.5}           & 0.9B  & 0.43    & 0.97  & 0.38   & 0.38     & 0.76   & 0.04     & 0.06  \\
SD-2.1~\cite{sd2.1}            & 0.9B  & 0.50    & 0.98  & 0.51 & 0.44     & 0.85   & 0.07     & 0.17 \\
Sana~\cite{sana}  & 1.6B  & 0.66  & 0.99  & 0.77   & 0.62 & 0.88  & 0.21  & 0.47 \\
LUMINA-Next~\cite{zhuo2024lumina}      & 2.0B  & 0.46    & 0.92  & 0.46 & 0.48 & 0.70 & 0.09 & 0.13  \\
SDXL~\cite{podell2023sdxl}             & 2.6B  & 0.55    & 0.98  & 0.74 & 0.39     & 0.85   & 0.15     & 0.23 \\
PlayGroundv2~\cite{li2024playground}   & 2.6B  & 0.59    & 0.98  & 0.73 & 0.67     & 0.82   & 0.14     & 0.22 \\
PlayGroundv2.5~\cite{li2024playground}   & 2.6B  & 0.56    & 0.98  & 0.77 & 0.52     & 0.84   & 0.11     & 0.17 \\
IF-XL~\citep{deepfloyd}            & 5.5B  & 0.61  & 0.97  & 0.74   & 0.66 & 0.81  & 0.13  & 0.35  \\

\midrule
Ours w/o KD    &  0.38B &     0.61   & 0.98   & 0.77   & 0.43  &  0.89    & 0.18   &  0.38 \\ 
SnapGen (ours)         &  0.38B & 0.66   & 1.00   & 0.84   & 0.60  &  0.88    & 0.18   &  0.45 \\ 
\bottomrule
\end{tabular}
\end{threeparttable}
}
\end{table*}

\begin{table}[!ht]
\caption{ \textbf{Detailed Results of DPG-Bench Comparisons.} }
\label{tab:dpgbench}
\centering
\resizebox{\linewidth}{!}{
\begin{tabular}{lc|ccccccc}
\toprule
Model       & Param & Overall $\uparrow$ & Global & Entity & Attribute & Relation & Other \\ 
\midrule
PixArt-$\alpha$~\cite{chen2023pixart}  & 0.6B  &  71.1 &  75.0 & 79.3  & 78.6  & 82.6  & 77.0  \\
PixArt-$\Sigma$~\cite{chen2024pixart}  & 0.6B  & 80.5 & 86.9  & 82.9  & 88.9  & 86.6 & 87.7 \\
SDv1.5~\cite{sd1.5}            & 0.9B  &  63.2 &  74.6 & 74.2  &  75.4 & 73.5  & 67.8 \\
SDv2.1~\cite{sd2.1}            & 0.9B  &  64.2 &  72.7 & 72.8  & 75.8  & 82.2  & 76.5  \\
Sana~\cite{sana} & 1.6B  & 84.8  & 86.0  & 91.5 & 88.9 & 91.9  & 90.7\\
LUMINA-Next~\cite{zhuo2024lumina}      & 2.0B  &  74.6 &  82.8 & 88.7  & 86.4  & 80.5  & 81.8  \\
SDXL~\cite{podell2023sdxl}             & 2.6B  &  74.7 & 83.3 & 82.4 & 80.9 & 86.8 & 80.4 \\
PlayGroundv2\cite{li2024playground}      & 2.6B  &  74.5 & 83.6 & 79.9 & 82.7 & 80.6 & 81.2 \\
PlayGroundv2.5\cite{li2024playground}    & 2.6B  &  75.5 & 83.1 & 82.6 & 81.2 & 84.1 & 83.5 \\
IF-XL~\citep{deepfloyd}            & 5.5B  &  75.6 & 77.7 & 81.2 & 83.3 & 81.8 & 82.9 \\
\midrule
Ours w/o KD      & 0.38B  &    76.3 & 77.8 & 83.7 & 84.3 & 86.7 & 77.4 \\ 
SnapGen (ours)             & 0.38B  &   81.1 & 88.3 & 85.1 & 87.0 & 87.3 & 87.6 \\ 
\bottomrule
\end{tabular}}
\end{table}

\section{Detailed Results on Benchmarks }\label{appendix:quantitative_benchmarks}

We present detailed results for GenEval and DPG-Bench in~\cref{tab:geneval} and~\cref{tab:dpgbench}, respectively. On GenEval, our model demonstrates exceptional performance in capturing color and positional attributes, with the counting subcategory showing significant improvement due to our proposed knowledge distillation (KD) scheme.

\section{Visualization of Few-step Generation }
\label{appendix:few_step}

\begin{figure*}[t]
    \centering
    \includegraphics[width=0.98\textwidth]{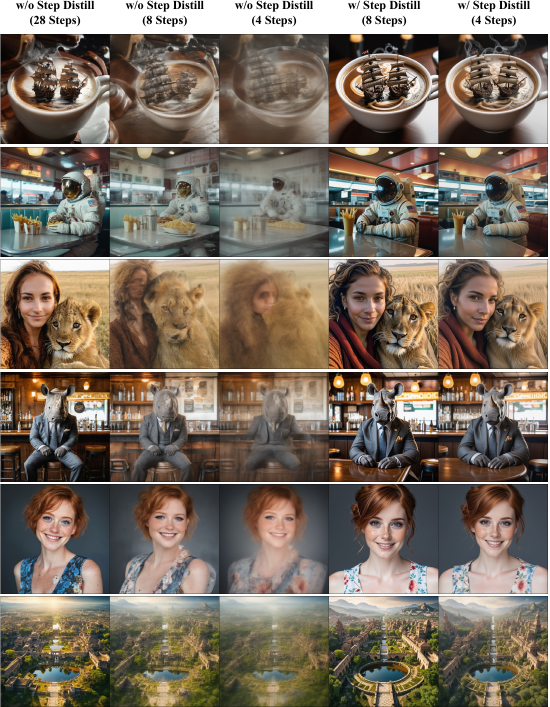}
    \caption{\textbf{Few-step generation qualitative comparison at $1024^2$ resolution.} The prompts used in these examples are from PixArt~\cite{chen2023pixart}.} \label{fig:fewstep-example}\vspace{-0.1in}
\end{figure*}

In~\cref{fig:fewstep-example}, we present qualitative comparisons of the 4- and 8-step T2I generation quality of our model, both with and without step distillation. The results demonstrate that the 4- and 8-step generations, following step distillation, not only significantly outperform the baseline model but also deliver quality comparable to the 28-step generation. Remarkably, the few-step generation captures finer details and mitigates the over-saturation issue commonly observed in the 28-step generation. Additionally, it exhibits superior prompt-following fidelity, making it a more efficient and effective approach for high-quality T2I generation.

\section{Additional T2I Comparison and Examples}
\begin{figure*}\vspace{-0.1in}
    \centering
    \captionsetup{type=figure}
    \includegraphics[width=\textwidth]{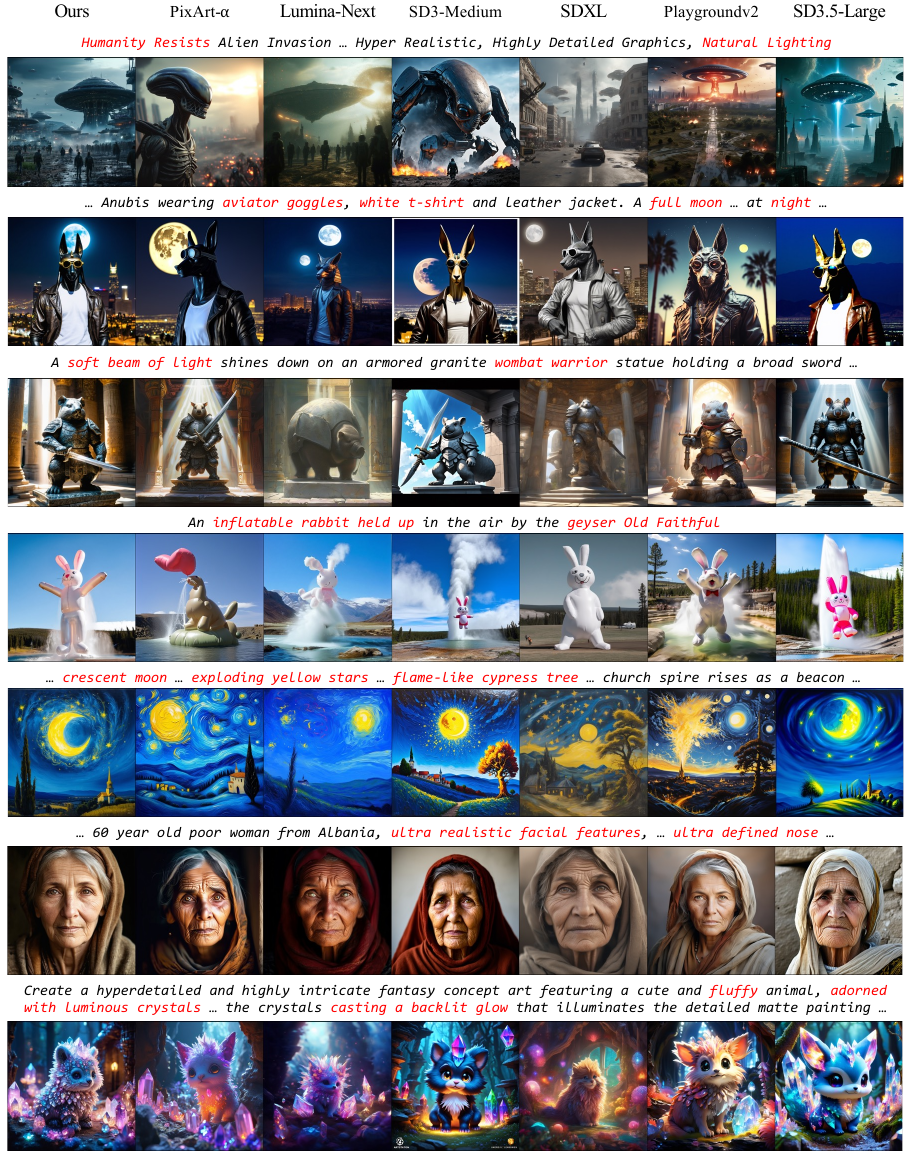}
    \vspace{-0.3in}
    \caption{\textbf{Additional Qualitative Comparison.} Our model demonstrates competitive visual quality and superior prompt-following ability.  Input text prompts are shown above each image grid; all images are generated at $1024^2$ resolution. Zoom in for details.}{
    }
    \label{fig:appendix_teaser1}
\end{figure*}

\begin{figure*}\vspace{-0.1in}
    \centering
    \captionsetup{type=figure}
    \includegraphics[width=\textwidth]{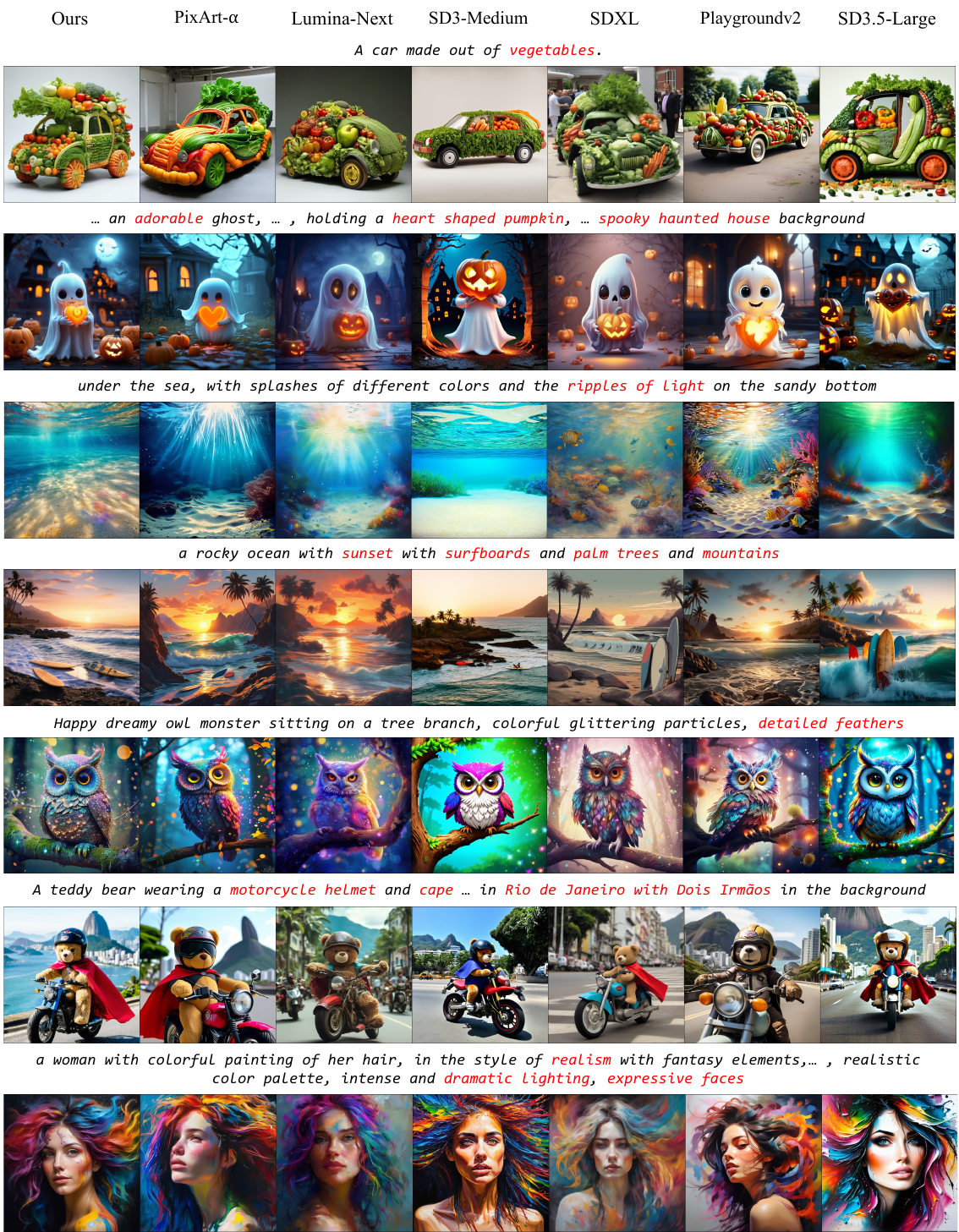}
    \vspace{-0.2in}
    \caption{\textbf{Additional qualitative comparison} at $1024^2$ resolution. Zoom in for details.}{
    }
    \label{fig:appendix_teaser2}
\end{figure*}

\begin{figure*}\vspace{-0.1in}
    \centering
    \captionsetup{type=figure}
    \includegraphics[width=\textwidth]{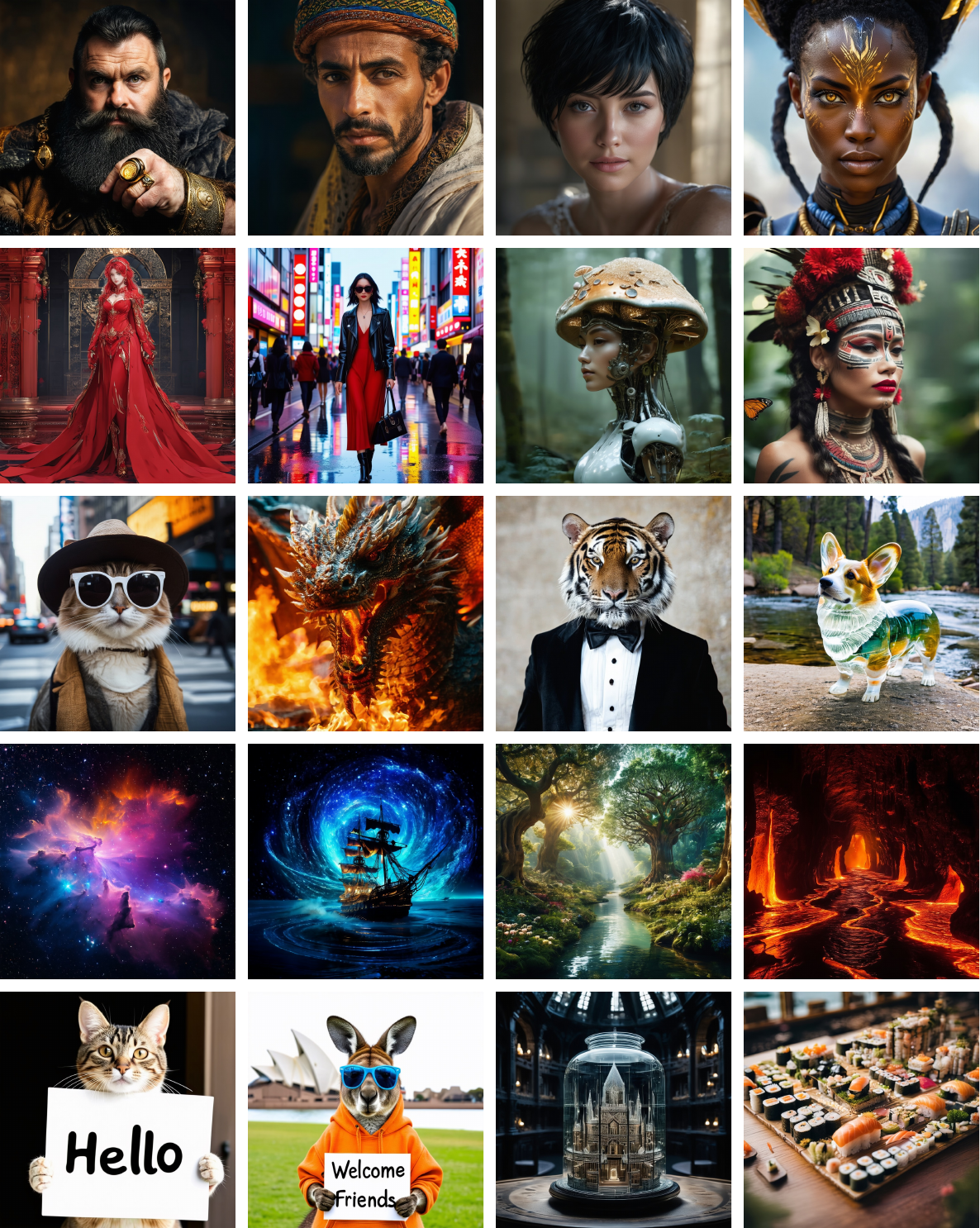}
    \vspace{-0.2in}
    \caption{\textbf{Additional T2I example visualization} at $1024^2$ resolution of our model. Zoom in for details.}{
    }
    \label{fig:appendix_vis1}
\end{figure*}

\begin{figure*}\vspace{-0.1in}
    \centering
    \captionsetup{type=figure}
    \includegraphics[width=\textwidth]{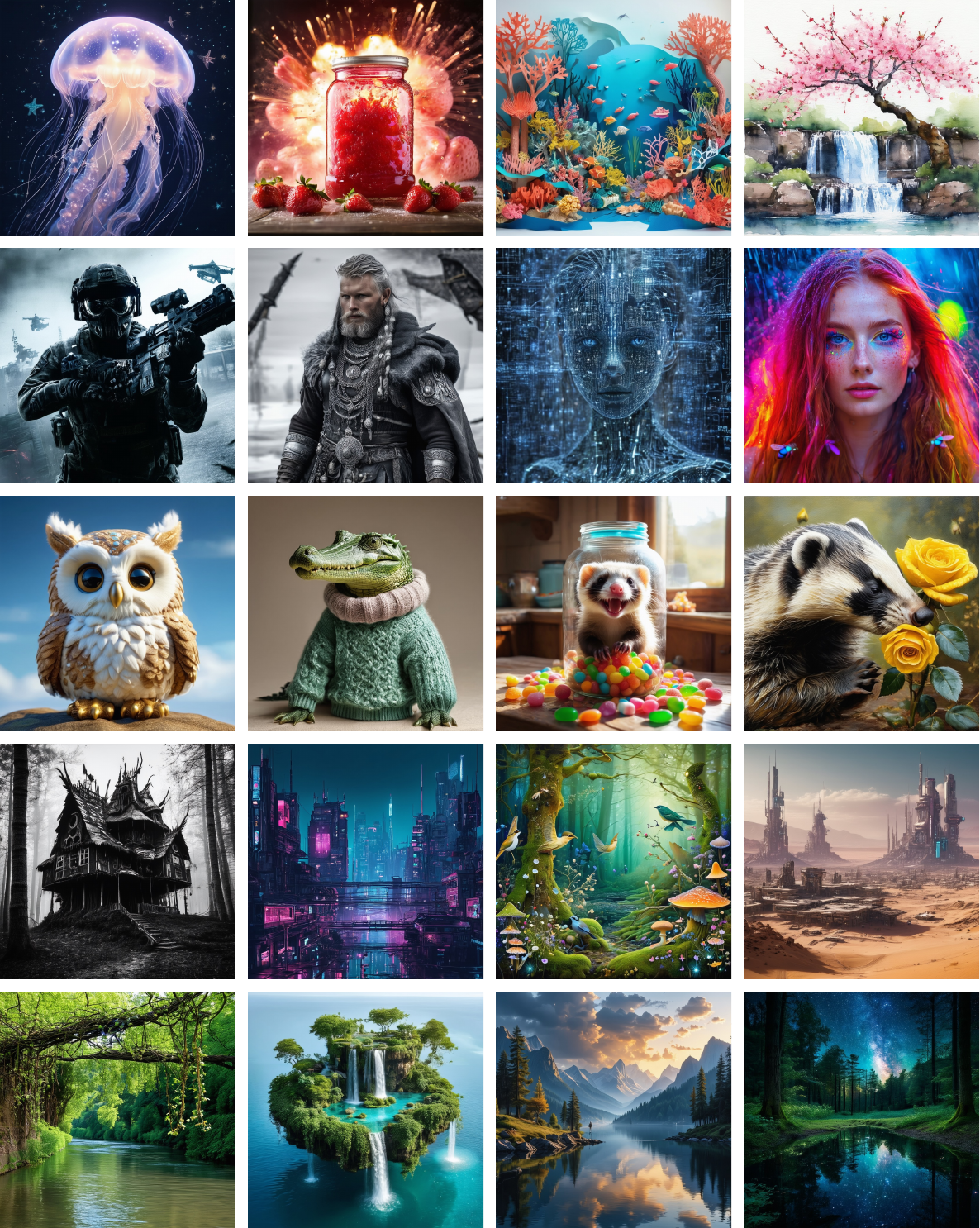}
    \vspace{-0.2in}
    \caption{\textbf{Additional T2I example visualization} at $1024^2$ resolution of our model. Zoom in for details.}{
    }
    \label{fig:appendix_vis2}
\end{figure*}

We present additional qualitative visualizations comparing 1024$\times$1024 generations across various T2I models in~\cref{fig:appendix_teaser1,fig:appendix_teaser2}. Furthermore, we showcase additional T2I examples generated by our model in~\cref{fig:appendix_vis1,fig:appendix_vis2}, with corresponding prompts detailed in~\cref{tab:prompt_for_vis}. These results demonstrate the exceptional prompt adherence and realistic generation quality achieved by our model.

\section{Ablation Study on Knowledge Distillation}\label{appendix:kd_ablation}

To demonstrate how the proposed knowledge distillation scheme improves the T2I generation quality of our model, we provide additional ablation studies into different distillation loss terms and the timestep-aware scaling operations ($t$-scaling) in Tab.~\ref{tab:appendix_kd_ablation}. As listed in the results, distillation at all levels consistently improves our model in both GenEval and ImageReward scores.

\begin{table}[H]
    \centering
    \caption{\textbf{Abalation Study on KD Components.}} 
    \vspace{-0.1in}
    \label{tab:appendix_kd_ablation}
    \resizebox{1.\linewidth}{!}{
    \begin{tabular}{cccc|cc}
        \toprule
        $\mathcal{L}_{\mathrm{task}}$ & $\mathcal{L}_{\mathrm{kd}}$ & $\mathcal{L}_{\mathrm{featkd}}$ & $t$-scaling & GenEval~$\uparrow$ & ImageReward~$\uparrow$ \\
        \midrule
          $\checkmark$ &  &  &  &  0.61 & 1.20\\
          $\checkmark$ & $\checkmark$ &  &  & 0.62 & 1.23 \\
          $\checkmark$ & $\checkmark$ & $\checkmark$ &  & 0.64 & 1.26 \\
          $\checkmark$ & $\checkmark$ & $\checkmark$ & $\checkmark$ & 0.66 & 1.32\\
        \bottomrule
        \end{tabular}
    }
\end{table}

\section{Large-scale T2I Training Details}
\label{appendix:additional_training_details}

Our training is conducted across 8 nodes, each equipped with 8 NVIDIA A100 80G GPUs, resulting in a total of 64 GPUs. The training batch size per GPU is set to 128 for 512$^2$ resolution, 48 for 1024$^2$ without knowledge distillation, and 32 for 1024$^2$ with knowledge distillation. Gradient checkpointing is employed to accommodate larger batch sizes. For step distillation, we use Fully Sharded Data Parallel (FSDP) and set gradient accumulation steps to 4, achieving an effective batch size of 16 per GPU.
We optimize the model using the AdamW optimizer with a weight decay of $0.01$ and $(\beta_1, \beta_2) = (0.9, 0.999)$. The learning rate remains constant during training and is scaled based on the batch size for the current training stage. For a total batch size of 1024, we use a learning rate of $5 \times 10^{-5}$.
The data collection and filtering pipeline for the T2I training dataset follows the approach described by \citet{kag2024ascan}. Additionally, we apply the Exponential Moving Average (EMA) with a decay rate of $0.999$.

\section{ImageNet-1K Class-conditional Generation}
\label{appendix:imagenet-1k}
We provide the experimental settings when examining each design choice in developing our efficient UNet. We train and evaluate them on the ImageNet-1K class-conditional generation task at a resolution of $256\times256$. To incorporate class conditions, we use a text template of the form ``a photo of \textless class name\textgreater'', which can be seamlessly fused by the cross-attention layer. As the dataset provides multiple names per class, we select one at random during both training and inference to enrich text mappings. This text is encoded by a compact CLIP-ViT/L14~\cite{radford2021learning} text encoder. 
For latent diffusion, we convert input images into latent using an 8-channel AE. We pre-compute both the image latent and the text embeddings, which reduces the GPU memory and non-trivial computation time during training. We adopt DDPM~\cite{ho2020denoising} as our training objective, applying a linear noise scheduler over $1000$ time steps. Models are trained for $120$ epochs (and $1000$ epochs for the final model) with a batch size of $1024$. The AdamW optimizer is used with a learning rate of $3e$-$4$, weight decay of $0.01$, and $(\beta_1, \beta_2) = (0.9, 0.99)$. For inference, we utilize the Heun~\cite{karras2022elucidating} discrete scheduler with $30$ sampling steps. We report the lowest FID score for each model variant, using classifier-free guidance (cfg)~\cite{cfg} within a scale range of $[1.3, 2.0]$. For the final model, we implement varied cfg~\cite{kynkaanniemi2024applying,chang2023muse} in steps $[10,30]$ with cfg scaling from 1.1 to 5.4.

\section{Evaluation Metrics Details}
\label{appendix:evaluation_metrics}

\noindent\textbf{GenEval}~\citep{ghosh2024geneval} is an object-focused evaluation framework for T2I models based on object detection and color classification to verify the fine-grained object properties in the generated images. Concretely, 6 tasks with different difficulties are focused in GenEval: single object, two object, counting, colors, position, and attribute binding. The prompts in the GenEval benchmark are generated from task-specific templates filled with randomly sampled object names (from 80 MS COCO~\cite{lin2014microsoft} class names), colors, numbers, and relative positions. There are a total of 553 prompts in GenEval and these prompts are usually concise (less than 20 tokens).

\noindent\textbf{DPG-Bench}~\citep{hu2024ella} is a benchmark mainly focused on dense prompts that describe multiple objects characterized by various attributes and relationships. The average number of tokens calculated by the CLIP tokenizer is 83.91, significantly longer than previous benchmarks such as 12.65 for T2I-CompBench and 12.20 for PartiPrompts. There are a total of 4286 prompts, spanning five categories: entity, global, attribute, relation, and other. 4 images are generated for each prompt and mPLUG-large~\cite{li2022mplug} is used as the adjudicator to evaluate the generated images according to the designated questions.

\noindent\textbf{Image Reward}~\citep{xu2024imagereward} is a zero-shot metric to encode the human preference on the text-to-image results. The model uses BLIP as the backbone and an MLP to obtain a scalar for preference comparison. After training on human-annotated preference data with ratings on alignment, fidelity, and harmlessness, the reward model aligns with human preference. Different from GenEval and DPG-Bench, we observe that Image Reward prefers images with detailed textures and backgrounds with rich color patterns.

\clearpage
\begin{table*}[h]
\caption{ \textbf{Prompts used for visualization.} }\label{tab:prompt_for_vis}
\vspace{-0.1in}
\begin{center}
\scriptsize
\begin{tcolorbox}[width=0.98\textwidth,title = {\textbf{Prompt used for Visualization in \cref{fig:appendix_vis1}, ($i$,$j$) represents the image in $i$-th row and $j$-th column}}]
(1,1) A black bearded dwarf with a big golden ring on his finger is looking seriously into the camera port \\
(1,2) Moroccan man, portrait, dynamic pose, detailed hair texture, ornate, sharp focus, in the style of National Geographic, photoportrait, Cinematic, ...\\
(1,3) beauty, short black hair, cinematic, photorealism, intricate ultra detail, high sharpness, 8K cinematic, photography, realistic, ...\\
(1,4) black African woman warrior character in the style of Avatar and Overwatch searching the path of true, extremely detailed skin, centered portrait, ...
\Sepline
(2,1) Lady in red, anime, cartoon, unreal engin, concept art, full body view, ornate, ultra detail, cinematic, beauty shot. \\
(2,2) A stylish woman walks down a Tokyo street filled with warm glowing neon and animated city signage. She wears a black leather jacket, a long red dress, and black boots, and carries a black purse. She wears sunglasses and red lipstick. She walks confidently and casually. The street is damp and reflective, creating a mirror effect of the colorful lights. Many pedestrians walk about. \\
(2,3) the mushroom on the head of a cyborg woman, in the style of atmospheric woodland imagery, hyperrealistic atmospheres, ... \\
(2.4) Close up out of focus blurry photo of a stunning interpretation of the most artistically and aesthetically refined representation of a Mayan tribal female models, heavy tribal makeup and facial tattoos with lush red lips, large traditional Mayan headdress, looking to the side, butterflies and flowers jungle background, minimal desaturated color palette, soft vignette, f1.4 110 sec shutter, soft light
\Sepline
(3,1) very realistic, detailed, cat with big hat and white sunglasses, posing, hyperrealistic, atmospheric, in city, street, new york, cinematic, dramatic lighting, photorealistic, Leica M10R 8k Leica 35mm lens, Tilt Blur, Shutter Speed 1 1000, F 5.6, Super Resolution \\
(3,2) a close-up of a fire spitting dragon, cinematic shot \\
(3,3) a tiger wearing a tuxedo \\
(3,4) Color photo of a corgi made of transparent glass, standing on the riverside in Yosemite National Park. 
\Sepline
(4,1) Colorful shining nebulae\\
(4,2) Pirate ship trapped in a cosmic maelstrom nebula, rendered in cosmic beach whirlpool engine, volumetric lighting, spectacular, ambient lights, light pollution, cinematic atmosphere, art nouveau style, illustration art artwork by SenseiJaye, intricate detail.\\
(4,3) A sureal parallel world where mankind avoid extinction by preserving nature, epic trees, water streams, various flowers, intricate details, rich colors, rich vegetation, cinematic, symmetrical, beautiful lighting, V-Ray render, sun rays, magical lights, photography\\
(4,4) River of lava flows through a hallway of caves
\Sepline
(5,1) a cat holds a sign saying "Hello"\\
(5,2) a portrait photo of a kangaroo wearing an orange hoodie and blue sunglasses standing on the grassin front of the Sydney Opera House holding a sign on the chest that says Welcome Friends \\
(5,3) Spectacular Tiny World in the Transparent Jar On the Table, interior of the Great Hall, Elaborate, Carved Architecture, Anatomy, Symetrical, ...\\
(5,4) tilt shift aerial photo of a cute city made of sushi on a wooden table in the evening.
\end{tcolorbox}
\vspace{0.2in}
\begin{tcolorbox}[width=0.98\textwidth,title = {\textbf{Prompt used for Visualization in \cref{fig:appendix_vis2}, ($i$,$j$) represents the image in $i$-th row and $j$-th column}}]
(1,1) aesthetic light colored blue jellyfish with stars\\
(1,2) realistic photo of a jar of strawberry jam with explosions\\
(1,3) A gorgeously rendered papercraft world of a coral reef, rife with colorful fish and sea creatures.\\
(1,4) A watercolor painting of a cherry blossom tree and waterfall
\Sepline
(2,1) call of duty ghosts \\
(2,2) black and grey, historical viking, Historic, historically accurate, black and grey clothes, with silver jewellery, full body, dark fantasy, hyperdetailed, intricate details, hyper realistic \\
(2,3) Create a virtual environment where the world has dissolved into a torrent of rapidly shifting code and floating in this place we see a beautiful female artificial intelligence whos skin, hair, and body are made out of patterns and sequences intertwining. The very essence of the virtual environment and the constructs within it are now exposed in this woman, revealing their true nature as complex algorithms and digital architecture. The womans algorithm poses a significant threat to her control over the virtual world\\
(2,4) portrait of pretty caucasian blueeyed woman with marked freckles on her cheeks, neon red flowing Hair, long hair with rainbow colors, neon Bright colors background, face makeup divided into 4 different parts with solid bright colors, colored light wasps fall like sparkles from rain in a romantic and glamorous atmosphere, the hair with an incredible movement that surrounds the whole scene, hyper realistic photography, 8k, high contrast in detail
\Sepline
(3,1) Pixar animation, little brown and white fluffy soft owl toy, sitting, ultra detailed, sky blue and golden details, 8k bright front lighting, fine luster, ... \\
(3,2) Crocodile in a sweater \\
(3,3) A mischievous ferret with a playful grin squeezes itself into a large glass jar, surrounded by colorful candy. The jar sits on a wooden table in a cozy kitchen, and warm sunlight filters through a nearby window \\
(3,4) A young badger delicately sniffing a yellow rose, richly textured oil painting.
\Sepline
(4,1) Monster Baba yaga house with in a forest, dark horror style, black and white.\\
(4,2) Midnight video game street with glitchy effects \\
(4,3) surreal magical fairy forest, soft brushstrokes, birds, dmt, fish, moss, wildflowers, mushrooms \\
(4,4) a cyberpunk city far away in a desert
\Sepline
(5,1) pretty river with overhanging vines \\
(5,2) A floating island with crystal-clear waterfalls and lush vegetation \\
(5,3) blue water lake reflecting clouds \\
(5,4) A deep forest clearing with a mirrored pond reflecting a galaxy-filled night sky
\end{tcolorbox}
\end{center}
\end{table*}
\clearpage